%% file: main.tex
\documentclass[runningheads]{llncs}

\input{preambles/packages}
\input{preambles/macro}

\begin{document}

\pagestyle{headings}
\mainmatter
\def\ECCVSubNumber{5500}  

\title{Style-Agnostic Reinforcement Learning} 

\titlerunning{Style-Agnostic Reinforcement Learning }

%
\author{Juyong Lee\orcidlink{0000-0002-8155-3998}\thanks{These authors contributed equally to this work.} \and
Seokjun Ahn\orcidlink{0000-0002-3769-9965}\repeatthanks \and   
Jaesik Park\orcidlink{0000-0001-5541-409X}}

\authorrunning{J. Lee et al.}

\institute{Pohang University of Science and Technology (POSTECH), South Korea\\
\email{\{joy.lee, sdeveloper, jaesik.park\}@postech.ac.kr}\\}
\maketitle

\input{sections/0_abstract}
\input{sections/1_introduction}

\input{sections/2_relatedwork}

\input{sections/3_backgrounds}
\input{sections/4_method}

\input{sections/5_experiments}
\input{sections/6_limitation}
\input{sections/7_conclusion}

\subsubsection*{Acknowledgement}
This work was supported by IITP grant funded by the Korea government(MSIT) (No.2019-0-01906, Artificial Intelligence Graduate School Program(POSTECH) and No.2022-0-00290, Visual Intelligence for Space-Time Understanding and Generation based on Multi-layered Visual Common Sense).


%
%
\bibliographystyle{splncs04}
\bibliography{egbib}

\clearpage

\renewcommand{\thefigure}{S\arabic{figure}}
\setcounter{figure}{0}

\renewcommand{\thetable}{S\arabic{table}}
\setcounter{table}{0}

\renewcommand{\thealgorithm}{S\arabic{algorithm}}
\setcounter{algorithm}{0}

\renewcommand{\theequation}{S\arabic{equation}}
\setcounter{equation}{0}

\makeatletter
\newcommand{\@chapapp}{\relax}%
\makeatother

\section*{\centering{Supplementary Material}}

\appendix
\begin{appendices}
\input{sections/8_appendix}

\end{appendices}

\end{document}

%% file: preambles/packages.tex
\usepackage{graphicx}
\usepackage{amsmath}
\usepackage{amssymb}
\usepackage{booktabs}
\usepackage{algorithm}
\usepackage{xcolor}
\usepackage{enumerate}
\usepackage{balance}
\usepackage{tikz}
\usepackage{comment}
\usepackage{color}
\usepackage{hyperref}
\usepackage{algpseudocode}
\usepackage{array}
\usepackage{multirow}
\usepackage{soul}
\usepackage{colortbl}
\usepackage{wrapfig}
\usepackage{orcidlink}
\usepackage[title]{appendix}

\definecolor{mustard}{rgb}{1.00,0.86,0.35}
\definecolor{corn}{rgb}{1.00,0.95,0.50}
\definecolor{parchment}{rgb}{1.00,1.00,0.76}

\newcolumntype{L}[1]{>{\raggedright\let\newline\\\arraybackslash\hspace{0pt}}m{#1}}
\newcolumntype{C}[1]{>{\centering\let\newline\\\arraybackslash\hspace{0pt}}m{#1}}
\newcolumntype{R}[1]{>{\raggedleft\let\newline\\\arraybackslash\hspace{0pt}}m{#1}}

\usepackage[capitalize]{cleveref}
\usepackage[accsupp]{axessibility}  

\graphicspath{{./figures/}}

%% file: preambles/macro.tex
\Crefname{sec}{Section}{Sections}
\Crefname{tab}{Table}{Tables}
\Crefname{eqn}{Equation}{Equations}

\newcommand{\etal}{{et al}.}

\usepackage{xr}
\makeatletter
\newcommand*{\addFileDependency}[1]{
  \typeout{(#1)}
  \@addtofilelist{#1}
  \IfFileExists{#1}{}{\typeout{No file #1.}}
}
\makeatother

\newcommand{\repeatthanks}{\textsuperscript{\thefootnote}}

\DeclareMathOperator{\E}{\mathbb{E}}


%% file: sections/0_abstract.tex
\begin{abstract}
We present a novel method of learning style-agnostic representation using both style transfer and adversarial learning in the reinforcement learning framework.
The style, here, refers to task-irrelevant details such as the color of the background in the images, where generalizing the learned policy across environments with different styles is still a challenge.
Focusing on learning style-agnostic representations, our method trains the actor with diverse image styles generated from an inherent adversarial style perturbation generator, which plays a min-max game between the actor and the generator, without demanding expert knowledge for data augmentation or additional class labels for adversarial training.
We verify that our method achieves competitive or better performances than the state-of-the-art approaches on Procgen and Distracting Control Suite benchmarks, and 
further investigate the features extracted from our model, showing that the model better captures the invariants and is less distracted by the shifted style.
The code is available at \href{https://github.com/POSTECH-CVLab/style-agnostic-RL}{https://github.com/POSTECH-CVLab/style-agnostic-RL}.
\keywords{Reinforcement Learning, Domain Generalization, Neural Style Transfer, Adversarial Learning}

\end{abstract}

%% file: sections/1_introduction.tex
\section{Introduction}

Learning visual representation in reinforcement learning (RL) framework incorporated with deep convolutional neural networks enabled achieving remarkable performances in various control tasks, including video games~\cite{Video1,Video2}, robot manipulation~\cite{Robot1,Robot2}, and autonomous driving~\cite{Navigation}. 
Unfortunately, however, generalization of the learned policies to unseen environments often results in failures, even with slight changes in the backgrounds~\cite{DGinRL1,DGinRL2,DGinRL3}.  

Several methods have been proposed to overcome this limitation of RL agents, such as having an encoder with generative models~\cite{DARLA,WorldModel,Dreamer,SLAC} or training with auxiliary tasks~\cite{CURL,CTRL,DARL}. Methods using generative models are designed to train the agents to understand the world environment, and auxiliary tasks enable the agent to extract better features that will lead to better performances. 
Due to its simplicity, the latter technique is gaining interest.
For example, recent works have shown that representation learning with self-supervision objectives~\cite{CURL,PAD}, data randomization with feature matching~\cite{RAND}, and data augmentation with additional regularization~\cite{DrQ,SODA,SVEA} result in high success.

The central concept of these approaches is to diversify training data so that the RL agents can learn invariants to the \emph{different styles of environments}. Here, the style of the environment indicates too detailed or irrelevant elements in the observation. In an autonomous driving situation, for instance, detecting the road or pedestrians is key to success, while the texture of the road, the colors of the other cars, or the weather condition can be regarded as different styles, which distract the agent from abstract and understand the situation.
Data augmentation, thus, might lead to better generalization capacity by mimicking natural style changes of observations. However, the results are inefficient or unstable without a careful choice of augmentation type and timing~\cite{RAD,InDA/ExDA}. To tackle this issue, sounder training methods of adding more regularization terms can be applied~\cite{DrQ,DrAC,SODA,SVEA}, but this makes the training objectives much more complex. 

In this work, we focus on learning style-agnostic representations and propose \textbf{SAR}: \textbf{S}tyle-\textbf{A}gnostic \textbf{R}L, which adopts the concept of both style transfer and adversarial learning.
Style transfer has been applied in many computer vision tasks, including domain generalization in RL~\cite{StyleGAN,StyleDG2,MixStyle}. 
Here, we further examine how style transfer is used to train the agents via generating images of new styles. 
The generator module in our model generates \emph{never-seen} styles and helps the actor generalize its learned policy to the unseen styles with various background images, including realistic images, without any heuristics or explicit environment class labels. 
Notably, the generator is trained with adversarial loss to perform adaptive style perturbation to the encoded feature representation.
To our best knowledge, this attempt and success have not been presented anywhere before.
An overview of our model is described in \autoref{fig:overview}.

In summary, the contributions of this paper are as follows:
\begin{itemize} 
\renewcommand{\labelitemi}{$\bullet$}
\item First, we introduce SAR, a novel method of learning style-agnostic representation for domain generalization in RL.
\item Second, we conduct extensive empirical evaluations showing that the model better captures invariants between different styles of environment.
\item Finally, we show that the SAR agents achieve competitive or better results on the Procgen~\cite{ProcGen} and Distracting Control Suite~\cite{DistCS} benchmarks than the previous state-of-the-art algorithms.
\end{itemize}

%% file: sections/2_relatedwork.tex
\section{Related work} 
\subsection{Domain Generalization in RL}
The main target of the domain generalization in RL can be summarized as training an agent to learn a robust policy that can be generalized to unseen environments. This allows RL algorithms to be applied in more realistic situations because agents are often tested in different environments from the training stage. One example is deploying a policy learned from the simulation to the real world in the robot manipulation task. 

Data randomization is a promising technique for such domain generalization in many cases~\cite{Robot2,Robot3}. However, it is difficult to build an accurate and practical simulator that enables using data randomization. Visual augmentation, on the other hand, is much easier to apply as it is based on simple image transformations. Laskin \& Lee \etal~\cite{RAD}, for example, demonstrated that simply using data augmentation, such as random cropping or gray scaling, is indeed helpful in improving the generalization capacity of RL agents. Also, Yarats \& Kostrikov \etal~\cite{DrQ} suggested using regularization terms for stabilizing the model training when using data augmentation. 

However, although data augmentation is potentially effective, it has several limitations. 
For example, 
a na\"ive choice of the augmentation type may degrade the generalization performance~\cite{RAD}. 
Applying cropping to an essential part of the image may confuse the agent, or training the model to produce the same action from a rotated image may be unreasonable. 
Here, we present a method for domain generalization by diversifying the training examples without requiring a complex strategy for data augmentation. The generator in the SAR model generates new feature examples having different styles and helps the agents with learning style-agnostic representations.

\subsection{Adversarial Feature Learning} 

Adversarial feature learning has become popular for domain generalization in computer vision tasks~\cite{SONG,ADV,MMDAAE,featureAttack,SagNet}. Li \etal~\cite{MMDAAE} showed that adversarial objectives help a model learn universal feature representations across different domains. Furthermore, Nam \& Lee \etal~\cite{SagNet} proposed a method of reducing the style gap for domain generalization in the image classification task. Inspired by this work, we investigate the adversarial feature learning for RL agents, but with a simpler training procedure, i.e., without dividing training phases or considering the environment style's \emph{class}es.

We note that adopting adversarial training for RL is not new~\cite{RARL,DARL}. 
To our best knowledge, however, exploiting adversarial learning to the latent features in RL framework and the min-max game scheme is not presented before.
Especially, our method can be interpreted as domain randomization beyond pixel space. 
Mixing styles with linear interpolation for representation learning in RL setting has been proposed in the earlier work~\cite{MixStyle}. However, unlike in the previous study, the style perturbation generator in SAR produces new synthetic styles that will not be seen with a simple interpolation. 
The adversarial examples help the actor extract style-agnostic embeddings without any label of styles and, finally, learn a robust policy for unseen environments.

%% file: sections/3_backgrounds.tex
\section{Backgrounds}

\subsection{Deep Reinforcement Learning}
RL agents interact and get trained with the world environment within a Markov decision process, which is defined as a tuple of $($state space $\mathcal{S}$, action space $\mathcal{A}$, transition probability $P$, reward space $\mathcal{R}$, and discount factor $\gamma)$; at every timestep $t$, the agent observes a state ${s_t} \in \mathcal{S}$ and takes an action ${a_t} \in \mathcal{A}$ from its policy $\pi({a_t}|{s_t})$~\cite{MDP}. Then, the agent is rewarded with $r_{t} \in \mathcal{R}$, and moves to the next state ${s_{t+1}}$ sampled from the transition probability $P(s_{t+1}|s_t, a_t)$. 

The policy of the agent is optimized to maximize the discounted sum of rewards $G_t = \sum_{k=t}^\infty \gamma^k r_k$. With given state $s_t$, the value of the state $V(s_t)$ is estimated as $\mathbb{E}_{\tau \sim \pi}[ G_t |s_t]$ and the value of the state-action $Q(s_t, a_t)$ is computed as $\mathbb{E}_{\tau \sim \pi}[ G_t |s_t, a_t]$, with trajectory $\tau$ sampled from the policy $\pi$.

With deep RL algorithms, the policy $\pi$ gets parameterized by a set of learnable parameters $\psi$, and value function $V$ or $Q$ is optimized with network parameter $\phi$. Also, especially for visual-based RL, since the images only offer partial observations, Mnih \etal~\cite{stackObs} has proposed that defining the state $s_t$ as a stacked consecutive image frames  $(o_{t-k}, o_{t-k+1}, \dots, o_t)$, where $\mathcal{O}$ is a high-dimensional image space and $o \in \mathcal{O}$, is effective.

\subsubsection{Proximal policy optimization} (PPO)~\cite{PPO} is a state-of-the-art on-policy RL algorithm that is used for, in our setting, discrete control tasks. Here, on-policy refers to a situation in which the model is trained with trajectories collected from the current policy. 
With PPO, the actor is updated using policy gradients, where the gradients are computed by using (i) action-advantages $A_t$ to reduce the gradient variances and (ii) clipped-ratio loss to constraint the update region. The critic estimates the state-value $V_\phi$, and gets trained with mean-squared error loss toward a target state-value $V_t^{target}$ using generalized advantage estimation \cite{PPO}. So, the objectives for the actor and critic network can be written as follows:
\begin{align}
    A_t &= Q_\phi(s_t, a_t) - V_\phi(s_t)\\
    \mathcal{L}_{\mathrm{actor}}(\psi) &= - \E_{s_t, a_t \sim \pi} \Big[ \min\Big( \frac{\pi_\psi(a_t|s_t)}{\pi_{\psi_{\mathrm{old}}}(a_t|s_t)} A_t, \mathrm{clip}\big( \frac{\pi_\psi(a_t|s_t)}{\pi_{\psi_{\mathrm{old}}}(a_t|s_t)}, \epsilon\big) A_t \Big)  \Big]\label{equation:ppo_actor}\\
    \mathcal{L}_{\mathrm{critic}}(\phi) &= \E_{s_t \sim \pi} \Big[ (V_\phi(s_t) - V_t^{target} )^2 \Big],\label{equation:ppo_critic}
\end{align}
where $\epsilon$ is a coefficient for clipping function $\mathrm{clip}(\cdot)\rightarrow[1-\epsilon,1+\epsilon]$.

\subsubsection{Soft actor-critic} (SAC)~\cite{SAC} is an off-policy RL algorithm for continuous control tasks. Since off-policy algorithms can train the agent with trajectories collected from the different policies, other than the current one, it appears to be more flexible to alternative routes but may get slower. With SAC, the actor learns a policy $\pi_\psi$, with the guide of critic estimating the state-action value $Q_\phi$ to maximize an objective as a sum of the reward and the policy entropy $\mathbb{E}_{s_t, a_t \sim \pi}[\sum_t r_t + \alpha {H}(\pi(a_t | s_t))]$. Here, $\alpha$ is an entropy coefficient determining the priority of exploration over exploitation.

The actor, then, is trained by maximizing the expected return of its sampled actions where the objective can be denoted as follows:
\begin{equation}
\label{equation:sac_actor}
    \mathcal{L}_{\mathrm{actor}}(\psi) = -\E_{a_t \sim \pi} \Big[Q_\phi(s_t, a_t) - \alpha \log \pi_\psi (a_t | s_t)\Big].
\end{equation}

The critic is updated to minimize the temporal difference. The objectives for the critic, with the estimated target value of the next state, are as follows:
\begin{align}
    V(s_{t+1}) &= \E_{a_t \sim \pi} \Big[ Q_{\phi}(s_{t+1}, a_t) - \alpha \log \pi_\psi (a_t | s_{t+1}) \Big]\label{equation:sac_value}\\
    \mathcal{L}_{\mathrm{critic}}(\phi) &= \E_{s_t, a_t, r_t, s_{t+1} \sim \mathcal{D}} \Big[\Big(Q_\phi (s_t, a_t) - \big(r_t + \gamma V(s_{t+1})\big) \Big)^2\Big]\label{equation:sac_critic}
\end{align}
where $\mathcal{D}$ is the replay buffer.

In this work, we show that our method can be attached to \emph{both on-policy and off-policy} RL algorithms, namely PPO and SAC. Also, our method can be applied to \emph{both continuous and discrete} control tasks as tested with the Procgen and Distracting Control Suite benchmark.
 
\subsubsection{Style transfer via instance normalization}
For style transfer, many recent works adopt a method of using instance normalization (IN)~\cite{IN,AdaIN,CIN,StyleTransfer1,MixStyle}. The underlying idea is that the mean and standard deviations of feature maps, computed across the spatial dimension within each feature channel, reflect the images' style. For example, the color or texture of an image can be captured with these statistics, which may be irrelevant features for classifying or detecting an object. By using IN, the effect of styles can be normalized with the formula:
\begin{equation}\label{equation:IN}
    \mathrm{IN}(z) = \gamma \cdot \frac{z - \mu(z)}{\sigma(z)} + \beta
\end{equation}
where $z \in \mathbb{R}^{C \times H \times W}$ is a feature map with channel $C$, height $H$ and width $W$, and $\beta, \gamma \in \mathbb{R}^C$ refers to the affine transformation parameters.

Note that $\mu(z) \in \mathbb{R}^C$ and $\sigma(z) \in \mathbb{R}^C$ are denoted as:
\begin{equation}\label{eqaution:IN_mu_sigma}
    \mu(z)_c = \frac{1}{HW} \sum_{h=1}^H \sum_{w=1}^W z_{c,h,w},~~~
    \sigma(z)_c = \sqrt{ \frac{1}{HW} \sum_{h=1}^H \sum_{w=1}^W (z_{c,h,w} - \mu(z)_c)^2 }
\end{equation}
with $c \in \{1, \dots, C\}$.

Moreover, Huang \& Belongie~\cite{AdaIN} proposed the method of adaptive instance normalization (AdaIN), which can be understood as replacing the style statistics of a target content image with those of a source style image with the definition below:
\begin{equation}\label{equation:AdaIN}
    \mathrm{AdaIN}(z, z') = \sigma(z') \cdot \frac{z - \mu(z)}{\sigma(z)} + \mu(z')
\end{equation}
where $z'$ is the feature map extracted from the source style image.

This idea can be used for mixing styles between images within a mini-batch. Especially in the domain adaptation for image classification, this has been proved to be successful~\cite{SagNet}. Zhou \etal~\cite{MixStyle} adopted style mixing for domain generalization in RL. However, the scope of mixing styles is restricted only to the training mini-batch as $\textrm{AdaIN}$ is an interpolation. Here, our method enables the agents to observe unseen styles by generating new adversarial feature examples. 

%% file: sections/4_method.tex
\begin{figure}[t]
\centering
\includegraphics[width=\textwidth]{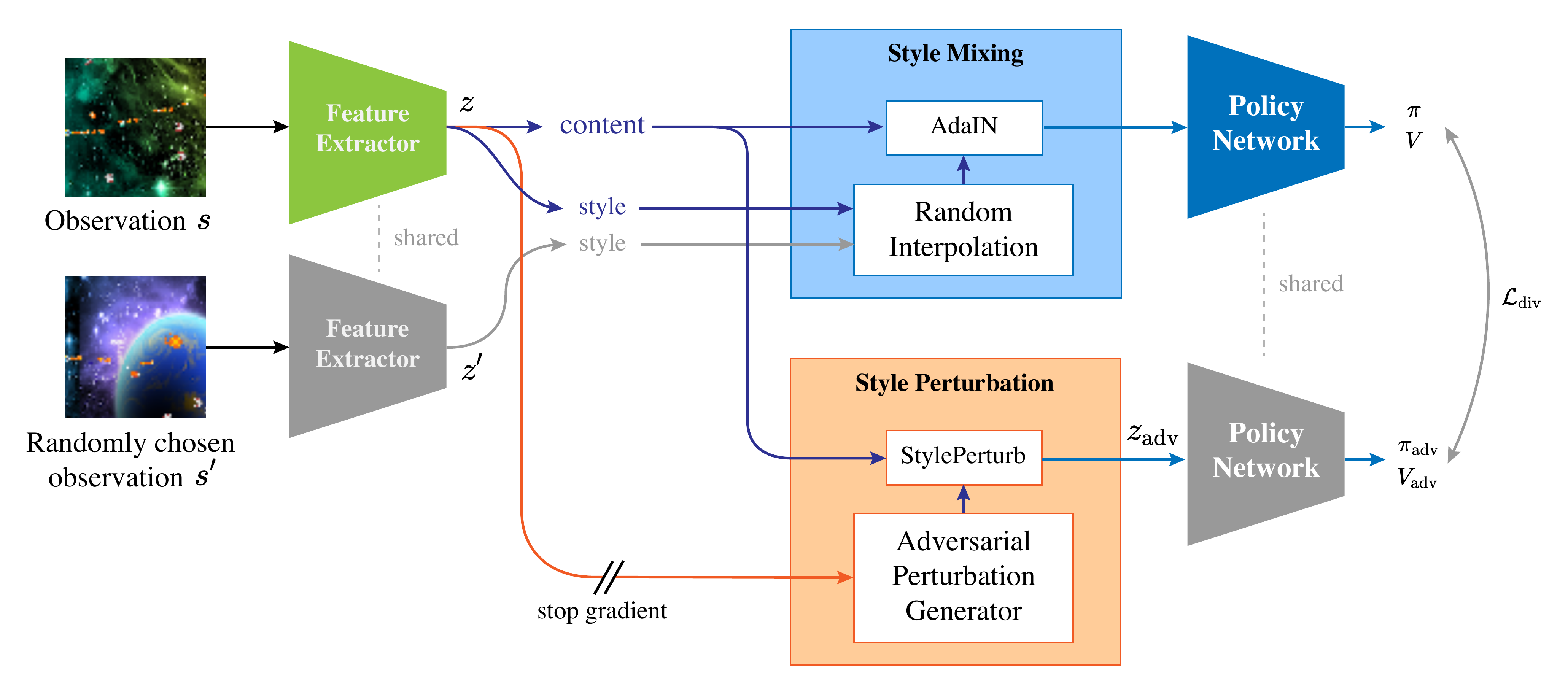}
\caption{Overview of the proposed Style-Agnostic Reinforcement learning (SAR) with the base model of PPO. The upper \emph{Style Mixing} module makes the policy network focus on the critical content in the observations by mixing styles from randomly chosen states $s'$. We newly employ our \emph{Style Perturbation} module, helping the agent with learning a robust policy by adversarially perturbing latent features.} 
\label{fig:overview}
\end{figure}

\section{Method}
\noindent\textbf{Overview.}
SAR is composed of an actor-critic module with RL objectives and a style perturbation generator helping the agents to observe more diverse styles of observations. While the generator is updated to produce more substantial perturbations for style transfer by \emph{maximizing} the difference between the action predictions, the actor learns a more robust policy to the attack from the generator by \emph{minimizing} the gap between predicted action distributions. 

To perform this min-max game between actor and generator, we present a \textbf{style perturbation layer}, shown in \autoref{fig:overview}. Unlike the conventional approach using only style mixing within the mini-batch~\cite{MixStyle}, the model in the training phase generates new styles and observes a broader range of feature examples. Note that this does not require explicit data augmentation that potentially degrades performance without a cautious choice of augmentation type.

\subsection{Style Perturbation Layer}
Our method is based on the concept of style transfer, which was proven to be successful in generating images with new styles \cite{CIN,StyleGAN}. The style perturbation layer shifts the style of observations $z$ with the generated perturbation mean  $\beta_{\mathrm{adv}}(z)$ and variance $\gamma_{\mathrm{adv}}(z)$, to build style-perturbed feature map $z_{\mathrm{adv}}$, or $\textrm{StylePerturb(z)}$, with the following equation:
\begin{equation}\label{equation:style_perturb}
    z_{\mathrm{adv}} = \gamma_{\mathrm{adv}}(z) \cdot \frac{z - \mu(z)}{\sigma(z)} + \beta_{\mathrm{adv}}(z).
\end{equation}

Then, the SAR agent should take the same action from $z_t$ and $z_{\mathrm{adv},t}$ to be robust among different environments, as the perturbed feature indicates an observation with different styles but the same semantics, e.g., in Procgen, the same player, enemies, and items, but shifted texture of the background image, the colors of projectiles, and the shapes of obstacles. We will further explain the objectives to achieve this generalization.

\subsection{SAR Objectives}
Primarily, the policy network is updated via PPO or SAC objectives. Thus, the actor loss of SAR is adopted from~\autoref{equation:ppo_actor} with PPO baseline or from~\autoref{equation:sac_actor} when using SAC. We will denote this loss be $\mathcal{L}^\circ_{\mathrm{actor}}$. Also, for the critic loss, as suggested in RAD~\cite{RAD}, we adopt the critic objective of PPO or SAC interchangeably, denoted as $\mathcal{L}^\circ_{\mathrm{critic}}$. 

Another big goal of SAR is to be robust to different environments. 
Therefore, the agent should learn its policy by minimizing the difference between the distributions of actions from the style-perturbed features $z_{\mathrm{adv},t}$ and the original ones $z_t$. By leveraging KL-divergence, we can calculate the objective as $\mathcal{L}_{\mathrm{div}}=\mathrm{KL}[\pi(\cdot|z_t) || \pi(\cdot|z_{\mathrm{adv},t})]$. Integrating this with a weight coefficient $\lambda$, the objective for the SAR actor module can be written as:
\begin{equation}\label{equation:sar_actor}
    \mathcal{L}_{\mathrm{actor}}(\psi) = \mathcal{L}^\circ_{\mathrm{actor}}(\psi) + \lambda \cdot \mathcal{L}_{\mathrm{div}}
\end{equation}

On the other hand, the generator participates in the min-max game in another manner: to maximize the differences between the action distributions. This module is trained with the objective of the same $\mathcal{L}_{\mathrm{div}}$ but with a converted sign. Unlike the previous works using class \emph{label} information of the environment style \cite{DARL} or additional heavy background images \cite{SVEA}, the objectives for the robust policy (i.e., adversarial loss) do not demand any secondary labors. Hence, the overall goals for the generator can be formalized as:
\begin{equation}\label{equation:sar_generator}
    \mathcal{L}_{\mathrm{gen}}(\theta) = - \lambda' \cdot \mathcal{L}_{\mathrm{div}},
\end{equation}
where $\lambda'$ can be different coefficient from that of actor objective.

Finally, the critic gets updated to guide the actor to optimize its policy to maximize the value function. Meanwhile, we observed that the sharing critic network, for predicting the value for both style-perturbed features and the original ones, does not bring a huge difference in the performance from decoupling the critic network but lighter training computation. Instead, we add a regularization term $G_{\mathrm{critic}}$ for the value function, to minimize the difference between the value predicted from the adversarial example, i.e., $ (V_\phi (z_t) - V_\phi( z_{\mathrm{adv},t}) )^2 $, which helps stabilization. Thus, the critic's objectives can be computed as follows:
\begin{equation}\label{equation:sar_critic}
    \mathcal{L}_{\mathrm{critic}}(\phi) = \mathcal{L}^\circ_{\mathrm{critic}}(\phi) + \kappa \cdot G_{\mathrm{critic}},
\end{equation}
with hyperparameter $\kappa$\footnote{The values used for each hyperparameters $\lambda$, $\lambda'$, $\kappa$ in the experiment are described in the supplementary material.}.

\subsubsection{On convergence.}
When the SAR agents learn the optimal policy $\pi^*$, the KL divergence term, or $\mathcal{L}_{\mathrm{div}}$, becomes zero. This is the situation where the actors infer the same actions from the features with different styles. This might be one of the two cases: (i) the generator produces the same style statistics for all images in the mini-batch, or more possibly, (ii) the actor well focuses on the invariant part of all observations.

Since the model should learn an additional generator module, the training procedure indeed demands more computations. However, the sample efficiency is not highly degraded even with limited training timesteps, e.g., the usual 25M timesteps in Procgen. Although the agent may not learn the optimal policy due to the limited number of epochs, we also empirically observed that the performances of the SAR agents converge as shown in \autoref{figure:curve}.

\subsection{Pseudo-code}
Here, we present the pseudo-code of the SAR algorithm.
As depicted in \autoref{fig:overview}, to maximize the effect of style transfer, we design the $z_t$ to pass a \emph{Style Mixing} module and a \emph{Style Perturbation} module with two divided branches. In the \emph{Style Mixing} module, the styles of observations in the mini-batch get interpolated with \autoref{equation:AdaIN}. In \emph{Style Perturbation} module, on the other hand, the styles of observations are shifted with new styles generated from the generator network with \autoref{equation:style_perturb}. 

With two different features $z_t$ and $z_{\mathrm{adv},t}$, the SAR agent predicts two different action distributions $\pi_t$ and $\pi_{\mathrm{adv},t}$. The difference between these predictions $\mathcal{L}_{\mathrm{div}}$ is computed, and it gets interpreted in two different ways: by the generator to produce more unfamiliar styles and by the actor to make its policy more robust.

\newcommand{\factorial}{\ensuremath{\mbox{\sc Factorial}}}
\begin{algorithm}
\caption{SAR algorithm}\label{euclid}
\begin{algorithmic}[1]
\State{Initialize rollout or replay buffer $\mathcal{D}$}
\State{Initialize parameters for policy $\psi$, generator $\theta$, and critic $\phi$}
\For{every epoch}
    \For{every environment step}
        \State{Sample $(s_t, a_t, r_t, s_{t+1})$}
        \State{Update $\mathcal{D} \leftarrow \mathcal{D}$ $\;\cup \; \{ (s_t, a_t, r_t, s_{t+1}) \}$}
    \EndFor
    \For{each mini-batch sampled from $\mathcal{D}$}
        \State{$z_t \leftarrow \mathrm{Encoder}(s_t)$} \Comment{Encoder in the actor network}
        \State{Generate $\beta_{\mathrm{adv}}(z_t), \gamma_{\mathrm{adv}}(z_t)$} \Comment{From the generator network}
        \State{$z_{\mathrm{adv},t} \leftarrow \mathrm{StylePerturb}(z_t)$}  \Comment{Use \autoref{equation:style_perturb}}
        \State{$z_t \leftarrow \mathrm{AdaIN}(z_t, z_t')$} \Comment{$z_t'$ is permuted from $z_t$ within mini-batch}
        \State{Compute $\mathcal{L}_{\mathrm{div}}$ from $z_t, z_{\mathrm{adv},t}$ }
        \State{Compute $\mathcal{L}_{\mathrm{actor}}$, $\mathcal{L}_{\mathrm{gen}}$, and $\mathcal{L}_{\mathrm{critic}}$ }
        \State{Update $\psi$, $\theta$, and $\phi$}
    \EndFor
\EndFor
\end{algorithmic}
\end{algorithm}

%% file: sections/5_experiments.tex
\section{Results}

\subsection{Setup}
In this section, we exhibit the experiment results for the generalization performance of our SAR model on Procgen~\cite{ProcGen} and Distracting Control Suite~\cite{DistCS} benchmarks. Recently, these benchmarks have become a standard for measuring the generalization performance of visual-based RL algorithms \cite{RAD,DrQ,DrAC,SODA,SVEA}. These contain reasonably challenging and diverse tasks, which are highly relevant to real-world robot learning.

\begin{wrapfigure}{R}{0.55\textwidth}
\centering
\includegraphics[width=0.55\textwidth]{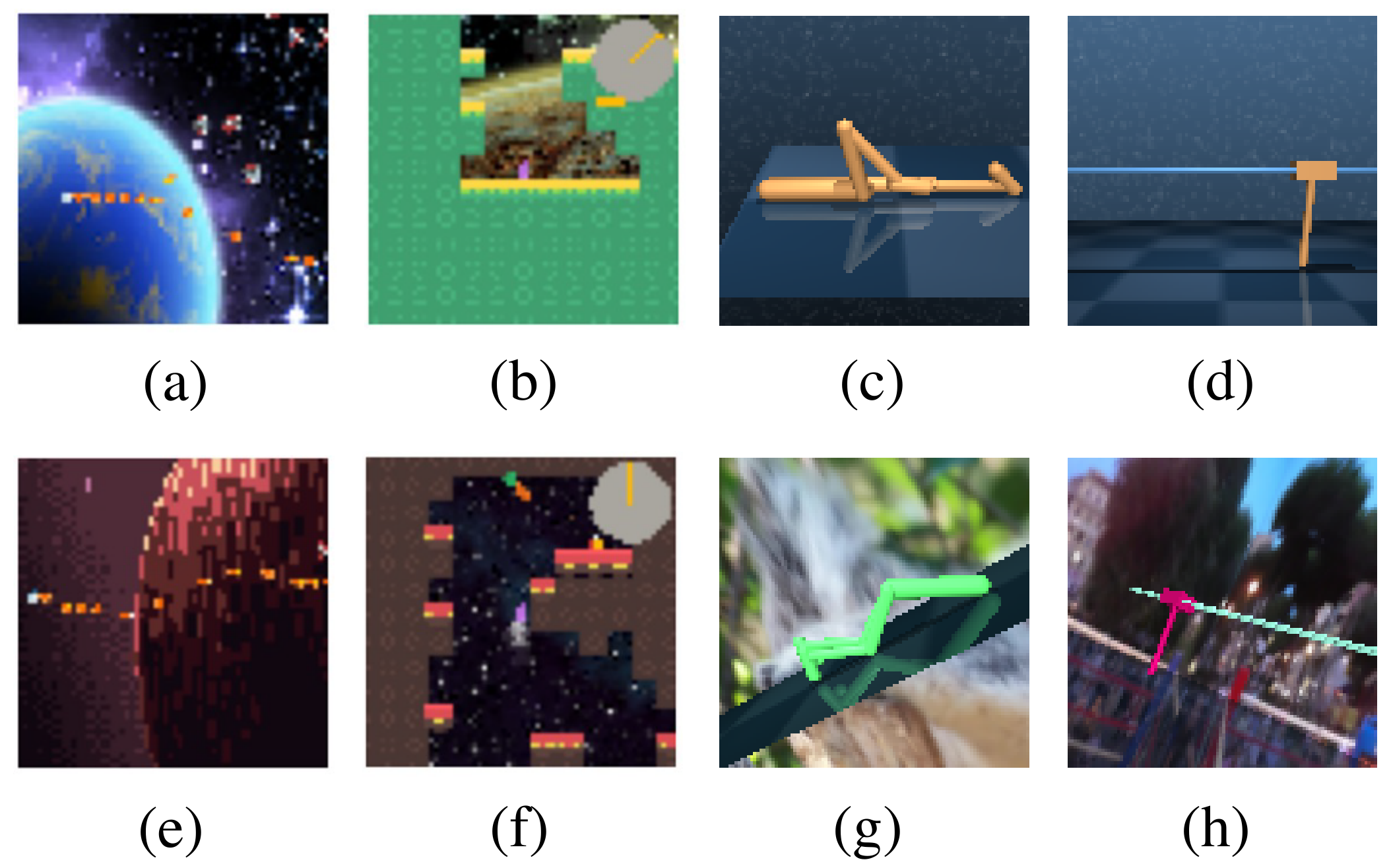}
\caption{Examples of seen training environments from (a) \texttt{starpilot} and (b) \texttt{jumper} in Procgen, (c) \texttt{walker:walk} and (d) \texttt{cartpole:balance} task in Distracting Control Suite, with examples of unseen test environments from (e) \texttt{starpilot} and (f) \texttt{jumper} in Procgen, (g) \texttt{walker:walk} and (h) \texttt{cartpole:balance} task in Distracting Control Suite.}
\label{figure:environments}
\end{wrapfigure}
 
While the Procgen benchmark is with a \emph{discrete} action space, the Distracting Control Suite presents \emph{continuous} control tasks. We exploited PPO as the basic baseline on the Procgen, and SAC as the basic baseline on the Distracting Control Suite, showing that the SAR algorithm can be applied to both on-policy and off-policy algorithms. \autoref{figure:environments} visualizes some examples of training and test environments in the two different benchmarks.

\noindent\textbf{OpenAI Procgen.} One key reason for choosing this benchmark is that this presents different styles between test and training environments. 
We train the agents on the first 200 levels in the Procgen environment. Then, we test the generalization performance of the agents on the environment levels sampled from the full distribution of unseen levels, with \emph{easy} distribution mode.
Among 16 tasks, we selected four tasks demonstrating comparably more considerable differences (\texttt{starpilot, climber, jumper, ninja}) and four tasks showing comparably less significant differences (\texttt{coinrun, maze, bigfish, dodgeball}) between the training and test environments style. 

\noindent\textbf{Distracting Control Suite.} DeepMind Control Suite~\cite{DMC} presents various continuous control tasks where RL agents can be tested. On top of the DMC, Stone \etal~\cite{DistCS} proposed Distracting Control Suite that distracts the agents by applying a color shift, changing the background images into videos, and rotating the camera angle. We test our model and other baselines with different noise coefficient values and show how these models generalize to unseen situations.

\subsection{Generalization Performance}
\noindent\textbf{Procgen.}
First, \autoref{table:procgen} shows the result of the generalization test of SAR with six other baselines. The SAR agent achieved high and robust performances in the zero-shot generalization test: 3 \emph{top-1 score}s and 7 \emph{top-3 score}s out of 8 tasks. 

The baselines are six visual-learning RL algorithms showing state-of-the-art results on Procgen. \textbf{PPO} \cite{PPO} is the vanilla on-policy RL baseline, and \textbf{RAD} \cite{RAD} uses data augmentation on top of PPO. We performed random translation (denoted as `trans') and random color cutout (denoted as `color') for RAD, as they are reporting the best performance. Among many advanced algorithms on RAD, \textbf{UCB DrAC}\cite{DrAC}, and \textbf{Meta DrAC} \cite{DrAC} are chosen to be compared with our method among three variants of DrAC; the former one presents the best performance among the variants. \textbf{Mixstyle} \cite{MixStyle} exploits the style mixing, and \textbf{DARL} \cite{DARL} uses an adversarial objective for regularization with style $ label $s.
\footnote{We reproduced all the results of the baselines. The results showed better than the reported performance in several tasks as more training steps~\cite{RAD,DrAC,MixStyle}. }

\begin{table}[h]
\centering
\caption{The generalization scores of SAR and baseline methods on Procgen. The results are averaged over three runs with 100M training timesteps {without smoothing}. The ranking stands for the average rank among all tasks. The \emph{top-1 score} is bold.}
\resizebox{\textwidth}{!}{
\begin{tabular}{l||C{1.5cm}C{1.5cm}C{1.5cm}C{1.5cm}C{1.5cm}C{1.5cm}C{1.5cm}|C{1.5cm}}
 & PPO \cite{PPO}                                                              & RAD \cite{RAD} (trans) 
 & RAD \cite{RAD} (color)
 & UCB DrAC~\cite{DrAC}
 & Meta DrAC~\cite{DrAC}
 & MixStyle \cite{MixStyle}                                                 
 & DARL \cite{DARL}                                                            & \textbf{SAR} (Ours) \\ 
\hline\hline
\texttt{Starpilot}        & {\begin{tabular}[c]{@{}c@{}}30.37\\±11.14\end{tabular}} & \begin{tabular}[c]{@{}c@{}}29.57\\±7.52\end{tabular}         &  {\begin{tabular}[c]{@{}c@{}}27.03\\±7.51\end{tabular}} & 
\cellcolor{parchment} {
\begin{tabular}[c]{@{}c@{}}33.17\\±6.37\end{tabular} }         & \begin{tabular}[c]{@{}c@{}}29.40\\±4.61\end{tabular}         & \begin{tabular}[c]{@{}c@{}}25.70\\±8.13\end{tabular}         & \begin{tabular}[c]{@{}c@{}}21.97\\±10.66\end{tabular}          & \cellcolor{corn}  \textbf{\begin{tabular}[c]{@{}c@{}} {{35.87}}\\±9.13\end{tabular}} \\ \hline

\texttt{Climber}          & \begin{tabular}[c]{@{}c@{}}6.73\\±1.27\end{tabular}            & \begin{tabular}[c]{@{}c@{}}4.87\\±1.31\end{tabular}          &   {\begin{tabular}[c]{@{}c@{}}7.23\\±2.05\end{tabular}}  & \cellcolor{corn} \textbf{\begin{tabular}[c]{@{}c@{}}9.43\\±1.35\end{tabular}}  & \begin{tabular}[c]{@{}c@{}}7.77\\±0.68\end{tabular}          & \begin{tabular}[c]{@{}c@{}}7.37\\±2.71\end{tabular}          & \begin{tabular}[c]{@{}c@{}}7.03\\±1.37\end{tabular}            & \cellcolor{parchment}  \begin{tabular}[c]{@{}c@{}}{7.93}\\±1.10\end{tabular}  \\ \hline

\texttt{Jumper}           &   {\begin{tabular}[c]{@{}c@{}}6.00\\±2.65\end{tabular}}            & \begin{tabular}[c]{@{}c@{}}4.67\\±0.58\end{tabular}          & {\begin{tabular}[c]{@{}c@{}}{5.67}\\±1.53\end{tabular}}  & {\begin{tabular}[c]{@{}c@{}}5.67\\±0.58\end{tabular}}  & 
\cellcolor{parchment} {\begin{tabular}[c]{@{}c@{}}7.33\\±2.52\end{tabular}}  &  \begin{tabular}[c]{@{}c@{}}6.00\\±2.65\end{tabular} &  \cellcolor{corn} \textbf{{\begin{tabular}[c]{@{}c@{}}7.67\\±1.53\end{tabular}}}   & \begin{tabular}[c]{@{}c@{}}{6.33}\\±1.15\end{tabular}  \\ \hline

\texttt{Ninja}            &  {\begin{tabular}[c]{@{}c@{}}6.00\\±2.83\end{tabular}}   & \begin{tabular}[c]{@{}c@{}}5.33\\±2.52\end{tabular}          &  \begin{tabular}[c]{@{}c@{}}5.33\\±2.08\end{tabular}           & \begin{tabular}[c]{@{}c@{}}6.33\\±1.53\end{tabular}           & \begin{tabular}[c]{@{}c@{}}7.33\\±0.58\end{tabular}           & \cellcolor{corn}  \textbf{\begin{tabular}[c]{@{}c@{}}{8.67}\\±1.53\end{tabular}} & \begin{tabular}[c]{@{}c@{}}7.33\\±0.58\end{tabular}            &  \cellcolor{parchment} {\begin{tabular}[c]{@{}c@{}}8.33\\±1.15\end{tabular}}  \\ \hline

\texttt{Coinrun}          &  {\begin{tabular}[c]{@{}c@{}} 8.67\\±1.15\end{tabular}}   & \begin{tabular}[c]{@{}c@{}}8.33\\±1.15\end{tabular}          & \cellcolor{corn}  \textbf{\begin{tabular}[c]{@{}c@{}} {9.33}\\±1.15\end{tabular}} &
\cellcolor{parchment}
\begin{tabular}[c]{@{}c@{}}9.00\\±1.00\end{tabular}           &   {\begin{tabular}[c]{@{}c@{}}8.33\\±0.58\end{tabular}} & 
\cellcolor{corn}  \textbf{\begin{tabular}[c]{@{}c@{}} {9.33}\\±0.58\end{tabular}} &
\cellcolor{corn}  \textbf{\begin{tabular}[c]{@{}c@{}} {9.33}\\±1.15\end{tabular}} &
\cellcolor{parchment}
{\begin{tabular}[c]{@{}c@{}}9.00\\±1.00\end{tabular}}  \\ \hline

\texttt{Maze}             & \begin{tabular}[c]{@{}c@{}}4.67\\±0.58\end{tabular}   &
\cellcolor{parchment} \begin{tabular}[c]{@{}c@{}}5.33\\±1.53\end{tabular}          &
\cellcolor{parchment}
\begin{tabular}[c]{@{}c@{}}5.33\\±0.58\end{tabular}  & 
\cellcolor{corn} \textbf{\begin{tabular}[c]{@{}c@{}} {7.33}\\±1.53\end{tabular}}  &
{\begin{tabular}[c]{@{}c@{}}4.67\\±0.58\end{tabular}} &
\cellcolor{parchment} {\begin{tabular}[c]{@{}c@{}}5.33\\±0.58\end{tabular}} & \begin{tabular}[c]{@{}c@{}}3.67\\±1.15\end{tabular}   & {\begin{tabular}[c]{@{}c@{}}5.00\\±1.00\end{tabular}}  \\ \hline

\texttt{Bigfish}          & \begin{tabular}[c]{@{}c@{}}10.37\\±3.27\end{tabular}           & \begin{tabular}[c]{@{}c@{}}6.03\\±2.06\end{tabular}          & \begin{tabular}[c]{@{}c@{}}10.13\\±1.84\end{tabular}           &  {\begin{tabular}[c]{@{}c@{}}9.37\\±3.16\end{tabular}} &
\cellcolor{parchment}
{\begin{tabular}[c]{@{}c@{}}12.03\\±4.38\end{tabular}} & \begin{tabular}[c]{@{}c@{}}9.00\\±2.94\end{tabular}          &  {\begin{tabular}[c]{@{}c@{}} {9.07} \\±4.05\end{tabular}} &   \cellcolor{corn} \textbf{\begin{tabular}[c]{@{}c@{}}13.20\\±6.16\end{tabular}} \\ \hline

\texttt{Dodgeball}        & {\begin{tabular}[c]{@{}c@{}}4.13\\±1.75\end{tabular}}  &  \cellcolor{parchment} {\begin{tabular}[c]{@{}c@{}}4.93\\±1.53\end{tabular}} & \begin{tabular}[c]{@{}c@{}}3.20\\±1.56\end{tabular}           & \cellcolor{corn}  \textbf{\begin{tabular}[c]{@{}c@{}}{8.13}\\±1.33\end{tabular}}  & \begin{tabular}[c]{@{}c@{}}2.40\\±2.46\end{tabular}          & \begin{tabular}[c]{@{}c@{}}3.60\\±2.31\end{tabular}          & \begin{tabular}[c]{@{}c@{}}4.47\\±2.73\end{tabular}            & \begin{tabular}[c]{@{}c@{}}3.60\\±2.23\end{tabular}           \\ \hline\hline
Avg. Rank        &  4.9   & 5.8 & 4.8 &  \cellcolor{parchment} 3.1 & 4.5 & 3.9 & 4.5 & \cellcolor{corn} \textbf{\begin{tabular}[c]{@{}c@{}} 2.9 \end{tabular}}           \\ \hline
\end{tabular}}
\label{table:procgen}
\end{table}

\noindent\textbf{Distracting Control Suite.}
As \autoref{table:dmc} demonstrates, SAR again showed robust performances in selected four tasks in Distracting Control suite compared to the baselines. This experiment implies that our method can also be applied in continuous control tasks and is attachable to the off-policy RL algorithms.

In this experiment, we purposely tested different baselines from Procgen to compare SAR with various algorithms. \textbf{SAC}~\cite{SAC} is the vanilla off-policy RL algorithm, and \textbf{CURL}~\cite{CURL} uses a contrastive objective for representation learning on top of SAC. \textbf{DrQ}~\cite{DrQ} was chosen as the representative baseline using the data augmentation with additional regularization terms. \textbf{PAD}~\cite{PAD} adapts to a new test environment using self-supervision.\footnote{We reproduced all the results of the baselines and applied `trans' to DrQ. The results with zero noise well match the reported performances in most cases~\cite{SAC,CURL,DrQ}.}\footnote{The performance of PAD differs from the reported value because of the simultaneous application of natural video backgrounds, color noise, and camera angle noise.}

\begin{table}[t]
\centering
\setlength{\tabcolsep}{5pt}
\caption{The generalization results on Distracting Control Suite after training 500k timesteps. The models are evaluated in two distraction settings: \texttt{moderate} setup with the noise coefficient $\beta_\mathrm{cam}=\beta_\mathrm{rgb}=0.3$ and 60 background videos, and \texttt{hard} setup with the noise coefficient $\beta_\mathrm{cam}=\beta_\mathrm{rgb}=0.5$ and 60 background videos, where $\beta_\mathrm{cam}$ and $\beta_\mathrm{rgb}$ mean camera angle noise intensity and color noise intensity. The results are averaged over 3 runs with different seeds, and rank is calculated within distracted environments. }
\label{table:dmc}
\begin{tabular}{l|c||C{1.35cm}C{1.35cm}C{1.4cm}C{1.35cm}|C{1.35cm}}
    \hline
    \multicolumn{2}{c||}{}           & \multicolumn{1}{c}{\begin{tabular}[c]{@{}c@{}}SAC\\ \cite{SAC} \end{tabular}}
    & \multicolumn{1}{c}{\begin{tabular}[c]{@{}c@{}}CURL\\ \cite{CURL} \end{tabular}} 
    & \multicolumn{1}{c}{\begin{tabular}[c]{@{}c@{}}DrQ\\ \cite{DrQ} \end{tabular}} 
    & \multicolumn{1}{c|}{\begin{tabular}[c]{@{}c@{}}PAD\\ \cite{PAD} \end{tabular}}
    & SAR (Ours) 
    \\ \hline\hline
    \multirow{3}{*}{{\begin{tabular}[l]{@{}l@{}}\texttt{walker}\\ \texttt{:walk}\end{tabular}}}      & zero noise          & 373±89 & 828±99 &  \cellcolor{corn} \textbf{930±23} & \cellcolor{parchment} 838±47 & 325±57    \\
                                               & \texttt{moderate}  & 96±10  & 88±11  & \cellcolor{parchment}126±33 & 125±27 & \cellcolor{corn} \textbf{139±19}    \\
                                               & \texttt{hard}      & \cellcolor{parchment}85±8   & 57±7   & 80±11 &  71±8 & \cellcolor{corn} \textbf{112±15}    \\ \hline
    \multirow{3}{*}{{\begin{tabular}[l]{@{}l@{}}\texttt{cartpole}\\ \texttt{:balance}\end{tabular}}}  & zero noise          & \cellcolor{corn}\textbf{996±1}   & \cellcolor{parchment}995±3  &\cellcolor{corn} \textbf{996±3}  & 992±6 & 990±5  \\
                                               & \texttt{moderate}  & \cellcolor{parchment}262±20 & 215±57 & 246±15 & 236±17 & \cellcolor{corn}\textbf{266±26} \\
                                               & \texttt{hard}      & \cellcolor{parchment}251±12  & 216±62 & 240±26 & 238±22 & \cellcolor{corn}\textbf{261±17} \\ \hline
    \multirow{3}{*}{{\begin{tabular}[l]{@{}l@{}}\texttt{reacher}\\ \texttt{:easy}\end{tabular}}}     & zero noise          & 197±7   & \cellcolor{corn} \textbf{960±24} & \cellcolor{parchment}844±63 & 671±285 & 177±51 \\
                                               & \texttt{moderate}  & \cellcolor{parchment}88±11   & 79±11  & 83±10 & 75±19  & \cellcolor{corn}\textbf{98±13}  \\
                                               & \texttt{hard}      & 72±11   & 67±12  & \cellcolor{parchment}78±3 & 71±8 & \cellcolor{corn}\textbf{93±10}  \\ \hline
    \multirow{3}{*}{{\begin{tabular}[l]{@{}l@{}}\texttt{cheetah}\\ \texttt{:run}\end{tabular}}}      & zero noise          & \cellcolor{parchment}316±159 & 280±12 & \cellcolor{corn}\textbf{332±21} & 285±29 & 304±80 \\
                                                & \texttt{moderate}  & \cellcolor{corn}\textbf{55±10}   & 46±8   & 47±8  & \cellcolor{parchment} 49±5 & \cellcolor{parchment}49±11  \\
                                                & \texttt{hard}     & \cellcolor{corn}\textbf{53±15}   & 41±8   & 33±13 & 41±10  & \cellcolor{parchment}46±13  \\ \hline\hline
    \multicolumn{2}{c||}{Avg. Rank}                                   & \cellcolor{parchment}2.125    & 4.625   & 3.125 & 3.625 &\cellcolor{corn} \textbf{1.25} \\ \hline
    \end{tabular}%
\end{table}

\begin{figure}[ht]
\centering
\includegraphics[width=0.8\textwidth]{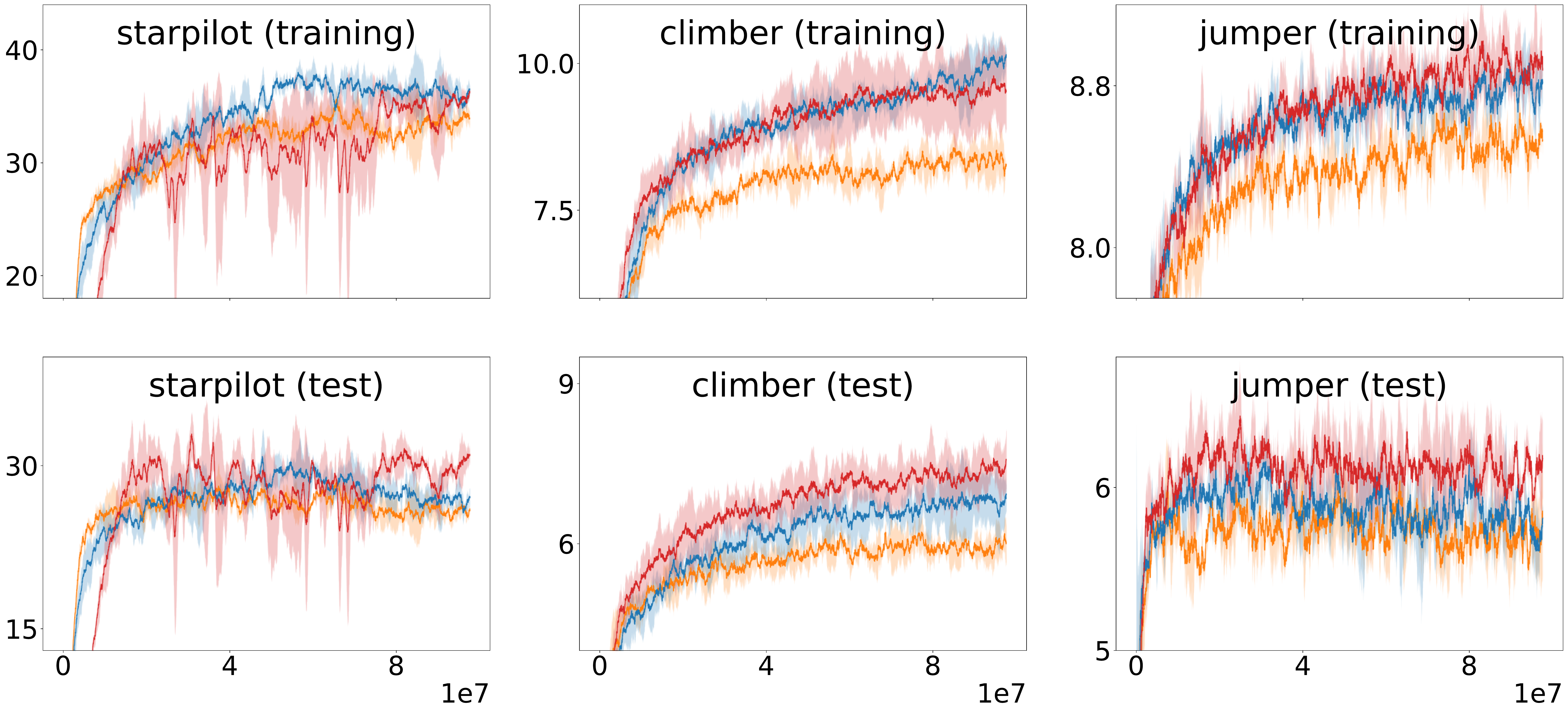}
\caption{
{
The learning curve of SAR with baselines. For better visualization, we selected three models and three tasks: PPO (blue), RAD (orange), and SAR (red). Here, we applied exponential moving average smoothing with a coefficient value of 0.98.
}
}
\label{figure:curve}
\end{figure}

\noindent\textbf{\\Model behavior.}
\autoref{figure:curve} provides the learning curve of the SAR agents. They exhibit competitive sample efficiency compared to the baselines. A quantitative comparison of the models' computational complexity is in \autoref{tab:complexity}. Although the SAR model requires more parameters, it does not sacrifice much training and test time in comparison with methods using data augmentation. 

\begin{table}[ht]
\centering
\setlength{\tabcolsep}{5pt}
\caption{A comparison between the number of parameters, training time, and test time. The training time refers to the time consumed for 256 timesteps and an update, and the test time is for running ten episodes in Procgen.}
\label{tab:complexity}
\begin{tabular}{l||ccccc|c}
\hline
 &
  \multicolumn{1}{c}{\begin{tabular}[c]{@{}c@{}}PPO\\ \cite{PPO} \end{tabular}} &
  \multicolumn{1}{c}{\begin{tabular}[c]{@{}c@{}}RAD~\cite{RAD}\\(color)\end{tabular}} &
  \multicolumn{1}{c}{\begin{tabular}[c]{@{}c@{}}UCB\\DrAC\cite{DrAC}\end{tabular}} &
  \multicolumn{1}{c}{\begin{tabular}[c]{@{}c@{}}MixStyle\\~\cite{MixStyle}\end{tabular}} &
  \multicolumn{1}{c|}{\begin{tabular}[c]{@{}c@{}}DARL\\~\cite{DARL}\end{tabular}} &
  \multicolumn{1}{c}{\begin{tabular}[c]{@{}c@{}}SAR\\(Ours)\end{tabular}} \\ \hline \hline
Parameters ($\times 10^6$)    & 0.626 & 0.626  & 0.626  & 0.626 & 0.678 & 1.151  \\ \hline
Training Time (s) & 6.507 & 11.605 & 12.841 & 6.735 & 6.542          & 13.377 \\ \hline
Test Time (s)     & 2.983 & 2.656  & 3.154  & 2.349 & 2.521          & 3.969  \\ \hline
\end{tabular}
\end{table}

\noindent\textbf{With augmentation.} Training the SAR agents can be integrated with other techniques. For example, \autoref{tab:additionals} presents the result of the SAR agents with data augmentation. Both the use of random translation and color cutout improved the performance. 
This result implies that the SAR agents can potentially be improved using other auxiliary tasks or regularization terms. 

\noindent\textbf{On curriculum learning.} 
{
Choice of timing for adopting the min-max game, i.e., curriculum learning, can improve final generalization performances for SAR.
See supplementary for the results of the experiment.
}

\begin{table}[h!]
\centering
\setlength{\tabcolsep}{5pt}
\caption{{
Results on generalization performances of SAR with the application of data augmentation and ablation study in \texttt{starpilot}. SAR ($\lambda=0$)  refers to the setting without adversarial loss, and SAR ($\kappa=0$) refers to the setting without regularization loss. We apply two different data augmentation: trans and color. The results are averaged over three runs. }
}
\label{tab:additionals}
\begin{tabular}{l||cc|ccccc}
\hline 
 &
  \multicolumn{1}{c}{\begin{tabular}[c]{@{}c@{}}PPO\\ \cite{PPO} \end{tabular}} &
  \multicolumn{1}{c|}{\begin{tabular}[c]{@{}c@{}}MixStyle \\ \cite{MixStyle} \end{tabular}} &
  \multicolumn{1}{c}{\begin{tabular}[c]{@{}c@{}}SAR\\($\lambda=0$)\end{tabular}} &
  \multicolumn{1}{c}{\begin{tabular}[c]{@{}c@{}}SAR\\($\kappa=0$)\end{tabular}} & 
  \multicolumn{1}{c}{\begin{tabular}[c]{@{}c@{}}SAR\\\end{tabular}} &
  \multicolumn{1}{c}{\begin{tabular}[c]{@{}c@{}}SAR\\(trans)\end{tabular}} &
  \multicolumn{1}{c}{\begin{tabular}[c]{@{}c@{}}SAR\\(color)\end{tabular}}
  \\ \hline
\texttt{starpilot}  & 
{\begin{tabular}[c]{@{}c@{}}27.09\\±0.83\end{tabular}} & 
{\begin{tabular}[c]{@{}c@{}}26.81\\±0.89\end{tabular}} &
{\begin{tabular}[c]{@{}c@{}}\textbf{27.44}\\±2.59\end{tabular}} &
{\begin{tabular}[c]{@{}c@{}}\textbf{29.28}\\±\textcolor{red}{7.79}\end{tabular}} &
{\begin{tabular}[c]{@{}c@{}}28.92\\±4.60\end{tabular}} &
{\begin{tabular}[c]{@{}c@{}}\textbf{30.76}\\±0.90\end{tabular}} &
{\begin{tabular}[c]{@{}c@{}}\textbf{33.72}\\±1.16\end{tabular}} \\ \hline 
\end{tabular}
\end{table}

\subsection{Ablation Study}
This ablation study answers two questions regarding (1) whether the generator module helps generalization performance and (2) generalization term $G_V$ is important for stabilization. Comparing SAR to PPO baseline, MixStyle using only style mixing, and SAR ($\lambda=0$), \autoref{tab:additionals} shows that the adversarial objective helps improve the mean of the performances. Comparing SAR to SAR ($\kappa=0$), \autoref{tab:additionals} shows that $G_V$ helps stabilize the variance of the performances. 
\footnote{Note that the results in \autoref{tab:additionals} are slightly different from \autoref{table:procgen}, as we applied exponential moving average smoothing before averaging with coefficient value 0.98.}

\subsection{Learned Feature Analysis}
Furthermore, we qualitatively examine the features extracted from the encoder learned with the SAR objectives. The feature $z$ we analyzed is from encoded features before entering the AdaIN layer to exclude the effect of explicit style mixing. 

We demonstrate three analyses on the embedding:
\begin{itemize} 
\renewcommand{\labelitemi}{$\bullet$}
\item GradCAM \cite{GradCAM} visualization for the high-level understanding interpretation.
\item Reconstruction images from the feature maps.
\item t-SNE \cite{t-SNE} for analyzing the latent representations.
\end{itemize} 

\subsubsection{Visualization of model decision}
\begin{figure}[h]
\centering
\includegraphics[width=0.99\textwidth]{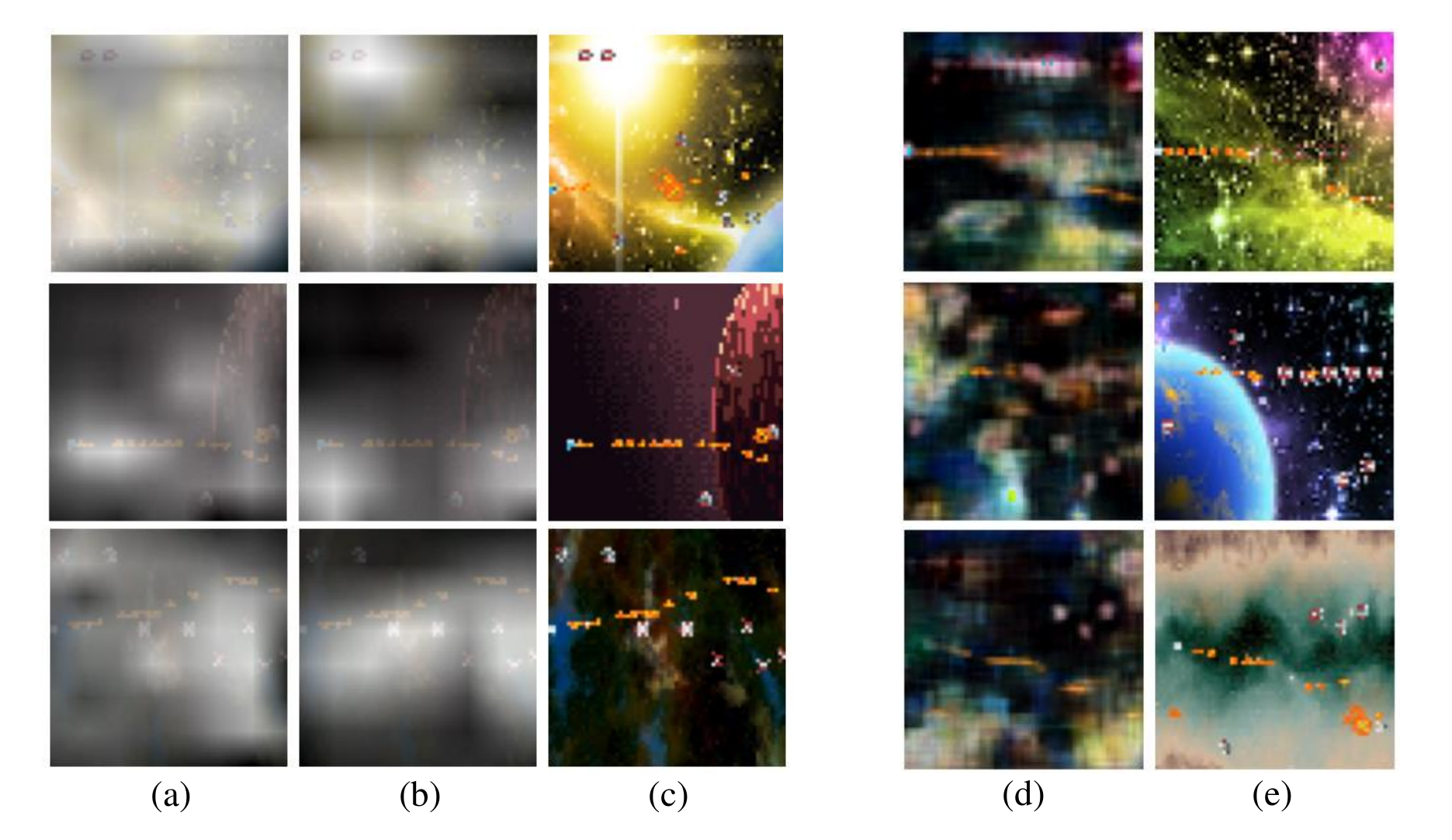}
\caption{GradCAM results of (a) PPO and (b) SAR, overlaid on (c) the original images from \texttt{starpilot} in Procgen. The highlighted regions represent where the agent is focusing. The SAR model better focuses on what is important with the style shifts. (d) Image reconstruction results from features extracted with the SAR agents, and (e) the original observations of \texttt{starpilot} in Procgen are displayed.}
\label{figure:frame_visualization}
\end{figure}

We use GradCAM \cite{GradCAM} to visualize where the trained agents are focusing with respect to the decisions.
GradCAM can be computed by averaging the activation scores across the channels of the target convolutional layer and weighting by their gradients. 
Both the agent trained by the vanilla PPO and our agents predict their actions as focused on similar objects in the training environment; in the case of \texttt{starpilot}, they are focusing on the shooter and the projectiles from enemies. In the unseen test environment shown in \autoref{figure:frame_visualization}, however, the vanilla PPO agent gets more distracted by the changed backgrounds and focuses on irrelevant areas in the images. 

\subsubsection{Image reconstruction from embedded features.}
Reconstructing images from the feature maps displays a more straightforward visualization of the characteristics of the learned features. We trained a new decoder network that converts the feature maps into the original images from training environments. 
In \autoref{figure:frame_visualization}, we show the reconstructed and original images. While the meaningful semantics, e.g., shooters or enemies, are remained, the reconstructed background seems invariant to the different original styles. 

\subsubsection{t-SNE Analysis.}
Li et al. \cite{DARL} addressed that the distance between the embedding in the latent space may reflect the dissimilarities between the features. Thus, by observing the t-SNE \cite{t-SNE} of the embedding from different environments, how the feature maps are correlated with the style of images can be visualized. While the features extracted from the PPO encoder are patterned with respect to the level \emph{label}s, i.e., the styles, the SAR encoder extracts invariant embedding regardless of them. The visualization result can be seen in supplementary materials.

%% file: sections/6_limitation.tex
\section{Limitation}
We address the limitation of the SAR agents, mainly shown in the noise-free setting in Distracting Control Suite in \autoref{table:dmc}, although they could well adapt to heavy noise.
The additional terms in learning objectives may negatively affect the performance when there is zero noise in the test environment. 
Not enough styles of training environments would have also affected the actors, as they could not observe a sufficient amount of styles of training features to compete well with the generator. 
The generator would have taken the wrong direction for generating the new styles, and a failure in the min-max game may happen.
Curriculum learning may help alleviate such concerns.


%% file: sections/7_conclusion.tex
\section{Conclusion}

The SAR agents learn style-agnostic representations by observing features with a wide range of styles by (i) mixing with style randomization and (ii) producing from an adversarial style perturbation generator. In both Procgen and Distracting Control Suite benchmark experimentation, the SAR agents show the best generalization performances in terms of rank. The qualitative analysis reveals that the model helps to learn style-agnostic representations. We hope that the progress made here provides a broader view bringing out more techniques for many other tasks as well, as the SAR agents do.

\vspace{-2mm}

%% file: sections/8_appendix.tex

\section{Implementation details}
We explain the implementation details for both Procgen~\cite{ProcGen} and Distracting Control Suite~\cite{DistCS} benchmark. We reproduce all the baseline results on top of implementation of PPO~\cite{PPOimpl} for Procgen and implementation of SAC~\cite{SACimpl} for Distracting Control Suite.

\subsection{Hyperparameters}
\subsubsection{Procgen.}
The baselines compared with our SAR model are PPO~\cite{PPO}, RAD~\cite{RAD}, UCB DrAC~\cite{DrAC}, Meta DrAC~\cite{DrAC}, MixStyle~\cite{MixStyle}, and DARL~\cite{DARL}. We follow the settings of Cobbe \etal~\cite{ProcGenImpl} in Procgen; the encoder in the actor-network is based on ResNet architecture~\cite{Impala}, and the encoded features are shared to both actor and critic networks. The encoder is composed of three layer-blocks, where one layer-block is built with five convolutional layers with two skip connections. The hyperparameters for the model and environments are well described in \autoref{table:hyperparam_procgen}.

For PPO, the basic baseline model, we use generalized advantage estimation~\cite{PPO} but no stacked observations~\cite{stackObs}. 

For RAD, we apply the random translation and color cutout, where their results are shown in \autoref{figure:aug}, as they are reported as the best~\cite{RAD}. 

For UCB DrAC and Meta DrAC, we follow the setting in~\cite{DrAC}. 

For MixStyle, the style mixing is done once after the feature passes through two layer-blocks, where the feature flow is divided into two branches in SAR. 

For DARL, we not only follow the adaptive coefficient in gradient reversal layer~\cite{DARL} but also control the effect of the domain adversarial loss with the coefficient of $d$ with the value reported in \autoref{table:hyperparam_procgen}. 

For SAR (Ours), we set the adversarial coefficients $\lambda, \lambda'$ to be equal, but they can be optionally different. 
We perform a grid search to find the best hyperparameter pairs and report them as indices (1)$\sim$(3) for the adversarial coefficient and (4)$\&$(5) for the value similarity coefficient, where each index refers to the task of:

\begin{itemize} 
\renewcommand{\labelitemi}{$\bullet$}
    \item (1) \texttt{starpilot}, \texttt{jumper}, \texttt{coinrun}
    \item (2) \texttt{climber}, \texttt{ninja}, \texttt{bigfish}
    \item (3) \texttt{maze}, \texttt{dodgeball} 
    \item (4) \texttt{starpilot}, \texttt{ninja}, \texttt{coinrun}
    \item (5) \texttt{climber}, \texttt{jumper}, \texttt{maze}, \texttt{bigfish}, \texttt{dodgeball}
\end{itemize}

\subsubsection{Distracting Control Suite}
We compare our SAR model with SAC~\cite{SAC}, CURL~\cite{CURL} and DrQ~\cite{DrQ} in Distracting Control Suite. The encoder network has 3 CNN layers with layer size 32 and kernel size 3. We set the stride of the first layer of the encoder to 2 and the stride of the remaining layers to 1. The encoder network is shared between actor and critic. In \autoref{table:hyperparam_dmc}, we report the value of the hyperparameters. 

For SAC and CURL, we reproduce the results based on the implementation in \cite{SACimpl} and \cite{CURL}.

For DrQ, we apply the random translation as they are reported as the best~\cite{DrQ}. We set the augmentation coefficients K=1 and M=1.

For SAR (Ours), we also set the adversarial coefficients $\lambda$, $\lambda'$ to be equal. We search for the best hyperparameter pair using grid search and report them as indices (1)$\&$(2) for the adversarial coefficient and (3)$\&$(4) for the value similarity coefficient, where each index refers to the task of:
\begin{itemize} 
\renewcommand{\labelitemi}{$\bullet$}
    \item (1) \texttt{walker:walk}, \texttt{cartpole:balance}
    \item (2) \texttt{reacher:easy}, \texttt{cheetah:run}
    \item (3) \texttt{walker:walk}, \texttt{reacher:easy}, \texttt{cheetah:run}
    \item (4) \texttt{cartpole:balance}
\end{itemize}

\subsection{Visualization of images}

\subsubsection{Augmentation result in Procgen}
\begin{figure}[h]
\centering
\includegraphics[width=0.45\textwidth]{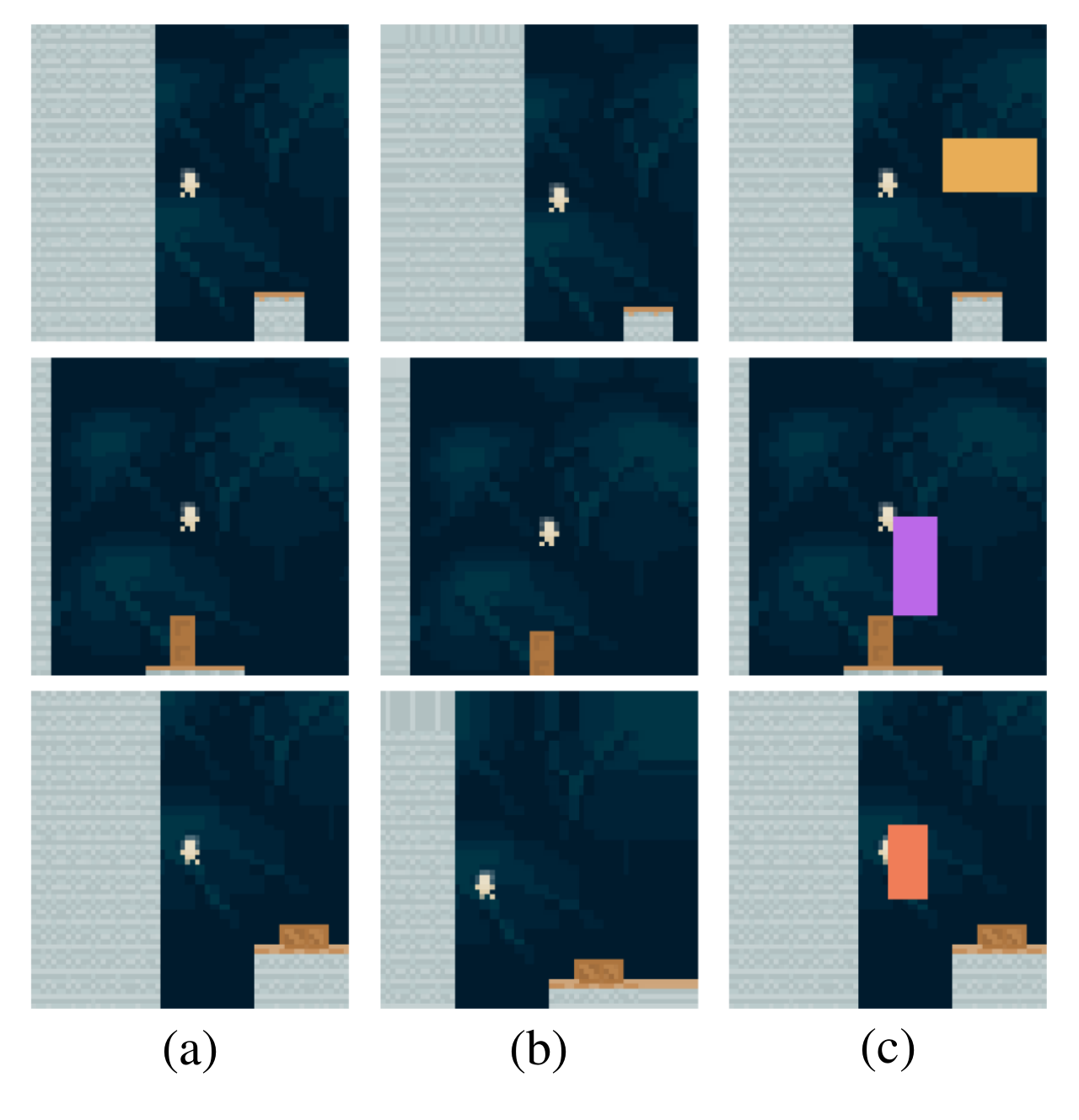}
\caption{(a) The original images, and augmentation results with (b) random translation and (c) random color cutout, of \texttt{coinrun} in Procgen. These methods are only applied to RAD in the generalization performance experiment. An additional experiment comparing SAR agents with and without these methods is conducted separately.}
\label{figure:aug}
\end{figure}

In our generalization performance experiment in Procgen, especially for RAD, we use two data augmentation methods: random translation and random color cutout. We show the results in \autoref{figure:aug}. 

\subsubsection{Distraction result in Distracting Control Suite.}
\begin{figure}[]
\centering
\includegraphics[width=0.95\textwidth]{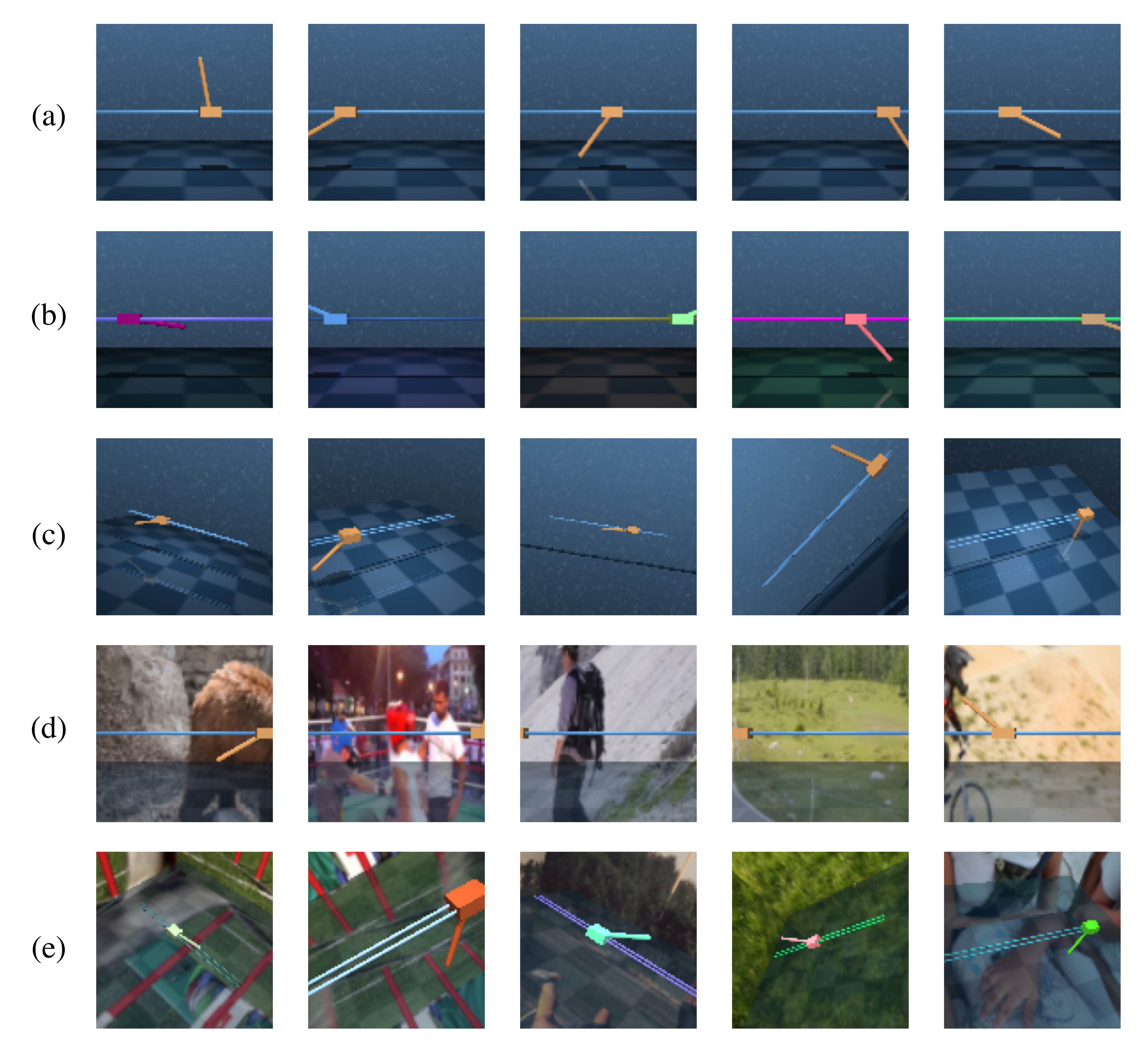}
\caption{Distracting environment examples of \texttt{cartpole:balance} task in Distracting Control Suite. Row (a) shows the task with zero noise. Row (b) shows the task with color shift $\beta_\textrm{rgb}=0.5$. Row (c) shows the with camera angle distraction $\beta_\textrm{cam}=0.5$. Row (d) shows the task with changed backgrounds. Row (e) shows the task with color shift $\beta_\textrm{rgb}=0.5$, camera angle distraction $\beta_\textrm{cam}=0.5$ and changed backgrounds.}
\label{figure:dmc_eval}
\end{figure}

In \autoref{figure:dmc_eval}, we visualize more diverse examples in Distracting Control Suite. The noises, referring to the distractions we apply, are shifts of color, distortions in the camera angle, and changing the background image into videos. Especially, the camera angle noise intensity, i.e., $\beta_{\textrm{cam}}$, in the main text refers to camera angle distraction intensity.

\pagebreak{}
\indent{}

\begin{table}[]
\centering
\begin{tabular}{L{5cm}L{2.7cm}}
\hline
\textbf{Hyperparameter}                                     & \textbf{Value}                                                                                                                                                                  \\ \hline \hline
Input image resolution                          & (64,64)                                                                                                                                                                  \\ \hline
Discount factor $\gamma$                          & 0.999                                                                                                                                                                  \\ \hline
Generalized advantage estimates                    & 0.95                                                                                                                                                                   \\ \hline
\# timesteps per rollout                           & 256                                                                                                                                                                    \\ \hline
\# epochs per rollout                              & 3                                                                                                                                                                      \\ \hline
\# minibatches per epoch                           & 8                                                                                                                                                                      \\ \hline
Entropy bonus                                      & 0.01                                                                                                                                                                   \\ \hline
PPO gradient clip range $\epsilon$                 & 0.2                                                                                                                                                                    \\ \hline
Reward normlization                                & yes                                                                                                                                                              \\ \hline                                                                         
Learning rate                                & 5e-4                                                                                                                                                       \\ \hline
\# workers                                         & 1                                                                                                                                                                      \\ \hline
\# environments per worker                         & 64                                                                                                                                                                     \\ \hline
\# total timesteps                                 & 100M                                                                                                                                                                   \\ \hline
Optimizer                                          & Adam                                                                                                                                                                   \\ \hline
Recurrent neural network                           & no                                                                                                                                                                     \\ \hline
Frame stack $k$                                      & no                                                                                                                                                                     \\ \hline
Regularization coefficient $\alpha_r$                                           & \begin{tabular}[c]{@{}l@{}}0.1 (UCB DrAC,\\Meta DrAC)\end{tabular}                                                                                                                                                                   \\ \hline
Exploration coefficient $c$                          & 0.1 (UCB DrAC)                                                                                                                                                                    \\ \hline
Sliding window size $K$                              & 10 (UCB DrAC)                                                                                                                                                                     \\ \hline
Domain loss coefficient $d$                              & 0.9 (DARL)                                                                                                                                                                     \\ \hline
Meta gradient clip range                              & 100 (Meta DrAC)                                                                                                                                                                     \\ \hline
Meta \# train steps                              & 1 (Meta DrAC)                                                                                                                                                                     \\ \hline
Meta \# test steps                              & 1 (Meta DrAC)                                                                                                                                                                     \\ \hline
Adversarial coefficient $\lambda$     & \begin{tabular}[c]{@{}l@{}}0.1 (1);\\0.01 (2);\\0.001 (3)\end{tabular} \\ \hline
Value similarity coefficient $\kappa$ & \begin{tabular}[c]{@{}l@{}}1.0 (4);\\0.1 (5)\end{tabular}                                                                             \\ \hline
\end{tabular}
\caption{Hyperparameters for SAR (Ours) and baselines in the Procgen experiment. The indices inside the parentheses for `adversarial coefficient' and `value similarity coefficient' indicate that the different values are used in different tasks.}
\label{table:hyperparam_procgen}
\end{table}

\begin{table}[]
\centering
\begin{tabular}{L{5cm}L{2.5cm}}
\hline
\textbf{Hyperparameter}                                     & \textbf{Value}                           \\ \hline \hline
Input image resolution                     & $(84, 84)$                                     \\ \hline
Discount factor $\gamma$                          & 0.99                           \\ \hline
Frame stack $k$                                    & 3                \\ \hline
Random shift                                      & Up to 4 pixels                \\ \hline
\begin{tabular}[c]{@{}l@{}}Action repeat\\\hspace{5mm} cartpole\\\hspace{5mm} finger\\\hspace{5mm} else\end{tabular}                                     & \begin{tabular}[c]{@{}l@{}}\\8\\2\\4\end{tabular}          \\ \hline
Episode length                             & 1000                               \\ \hline
Replay buffer size                          &  100000                              \\ \hline
Optimizer   & Adam          \\ \hline
\begin{tabular}[c]{@{}l@{}}Learning rate\\\hspace{5mm} actor, critic, attacker\\\hspace{5mm} alpha\end{tabular}                                     & \begin{tabular}[c]{@{}l@{}}\\$10^{-3}$\\$10^{-4}$\end{tabular}          \\ \hline
Encoder feature dimension                                     & 50                            \\ \hline
\begin{tabular}[c]{@{}l@{}}Target smoothing coefficient $\tau$\\\hspace{5mm} actor\\\hspace{5mm} critic\\\hspace{5mm} alpha\end{tabular}   & \begin{tabular}[c]{@{}l@{}}\\0.05\\0.01\\0.5\end{tabular}          \\ \hline
Target update interval                & 2                             \\ \hline
Batch size                & 128                             \\ \hline
Latent dimension                & 128                             \\ \hline
Initial temperature                & 0.1                             \\ \hline
Initial steps                & 1000                             \\ \hline
\begin{tabular}[c]{@{}l@{}}Network update frequency\\\hspace{5mm} attacker, critic\\\hspace{5mm} actor\end{tabular}   & \begin{tabular}[c]{@{}l@{}}\\1\\2\end{tabular}          \\ \hline
Adversarial coefficient $\lambda$     & \begin{tabular}[c]{@{}l@{}}0.01 (1);\\0.1 (2)\end{tabular} \\ \hline
Value similarity coefficient $\kappa$ & \begin{tabular}[c]{@{}l@{}}0.1 (3);\\1.0 (4)\end{tabular}                                                                             \\ \hline
\end{tabular}
\caption{Hyperparameters for SAR (Ours) and baselines in the Distracting Control Suite experiment.}
\label{table:hyperparam_dmc}
\end{table}

\pagebreak{}
\section{Learning curves}
We plot the learning curve of SAR agents and baselines in both Procgen and Distracting Control Suite. For clear visualization, each graph is illustrated with smoothing, following the settings of Cobbe \etal~\cite{ProcGenImpl}. 

\subsubsection{Procgen.} 
\begin{figure}[h]
\centering
\includegraphics[width=0.95\textwidth]{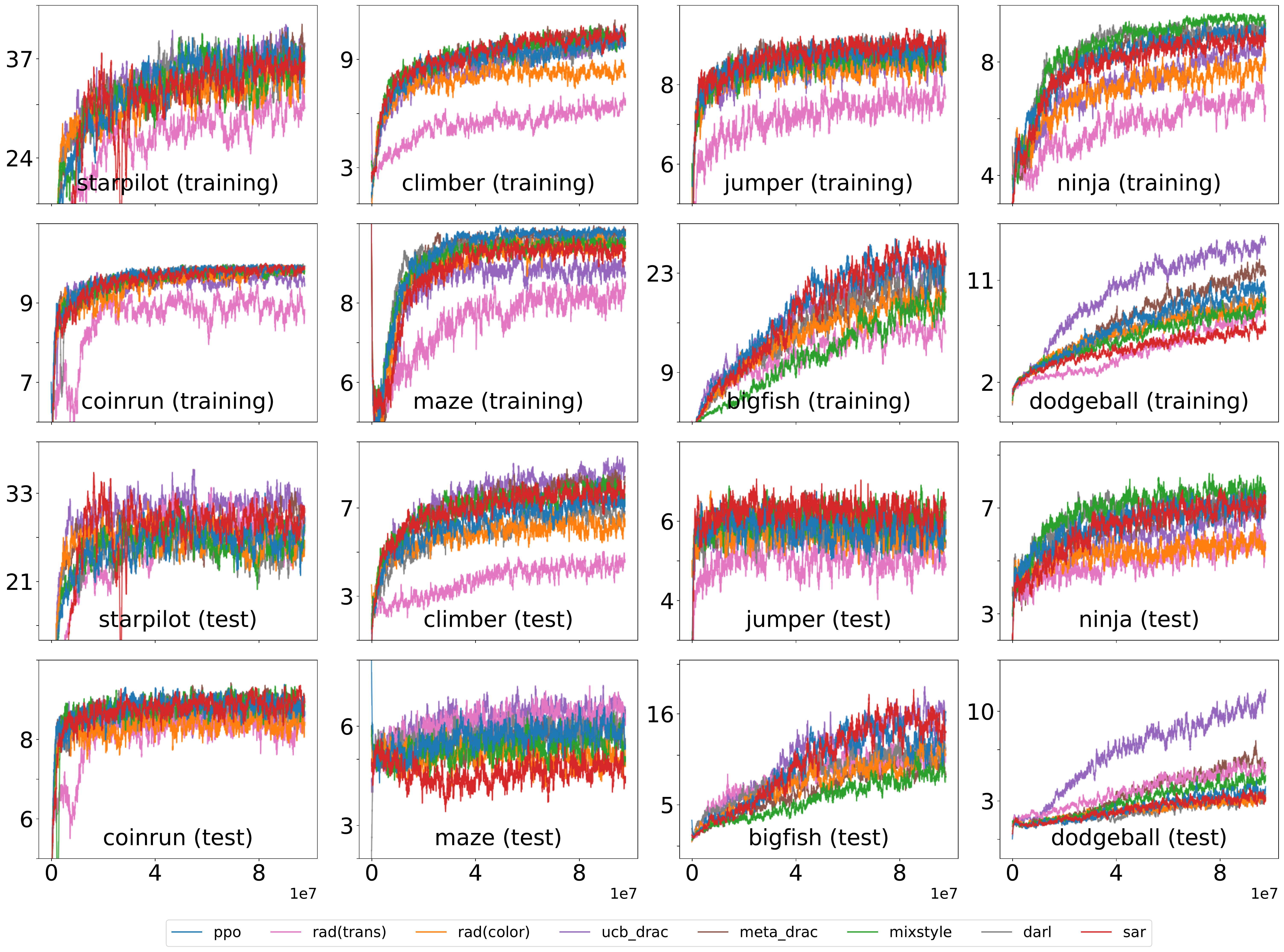}
\caption{
{
The learning curves of training and test in Procgen.
}}
\label{figure:procgen_training}
\end{figure}

\autoref{figure:procgen_training} shows the learning curves of models with the best performances from each algorithm in the Procgen environment. 
We apply an exponential moving average smoothing with the smoothing coefficient value of $0.95$.

\subsubsection{Distracting Control Suite.} 
\begin{figure}[h!]
\centering
\includegraphics[width=0.8\textwidth]{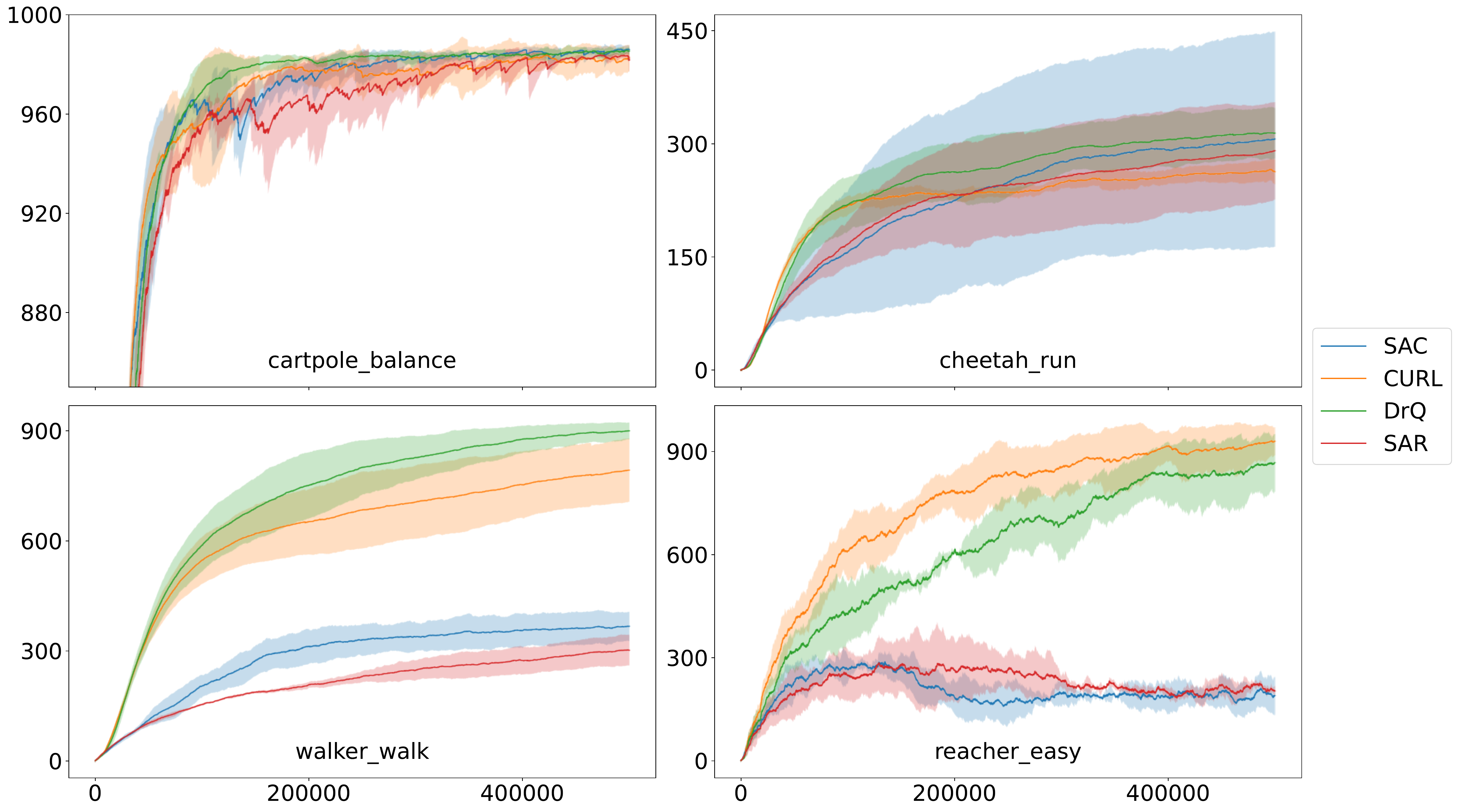}
\caption{The learning curves of training in Distracting Control Suite.}
\label{figure:dcs_training}
\end{figure}

\begin{figure}[h!]
\centering
\includegraphics[width=0.8\textwidth]{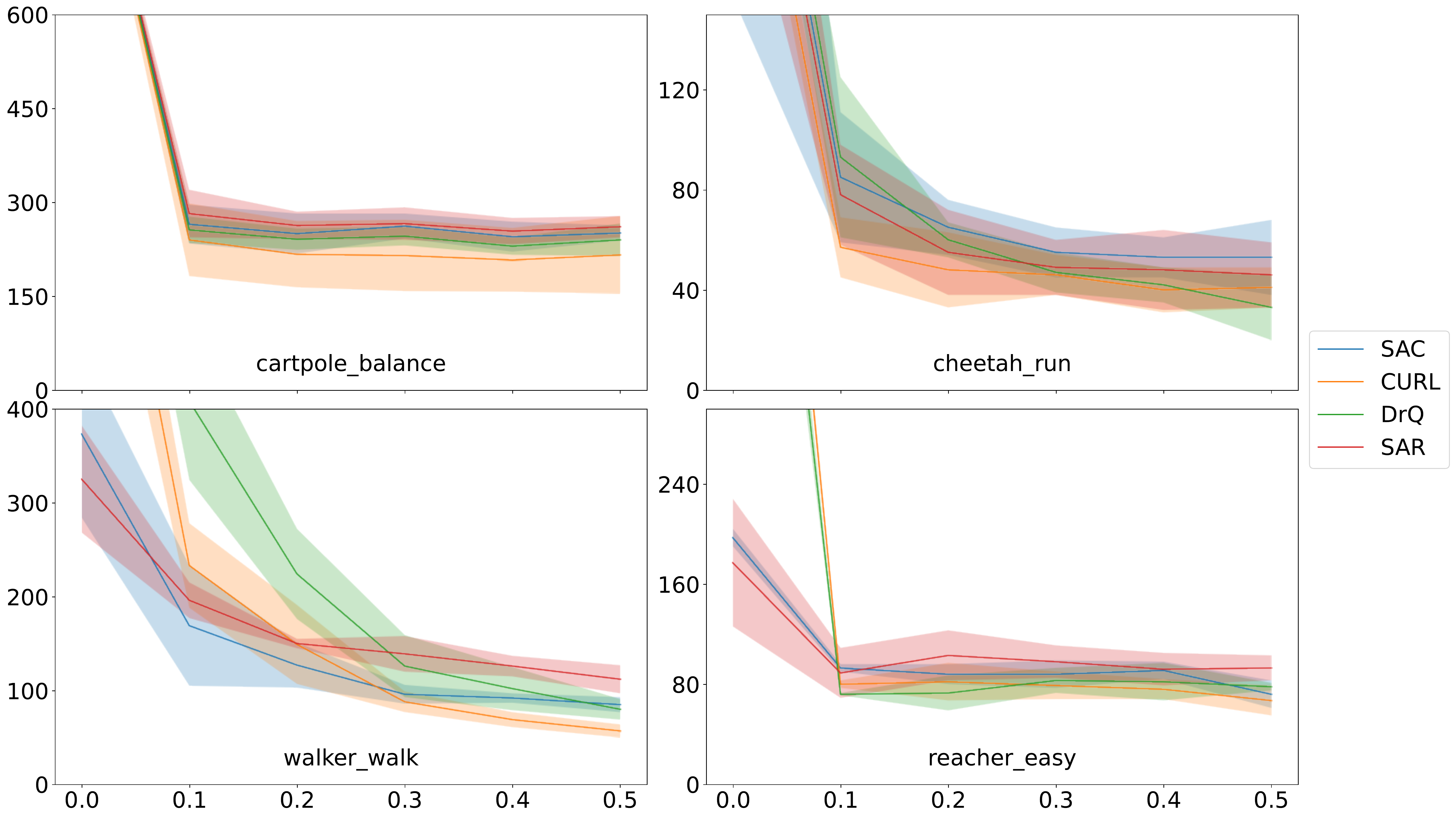}
\caption{Evaluation performances with respect to various distraction scales in Distracting Control Suite. $\beta_{\textrm{rgb}}$ and $\beta_{\textrm{cam}}$ are the same with x-axis distraction values. We use 4, 8 and 60 background videos in 0.1, 0.2 and $>$0.3 distraction values respectively. The SAR agents show better generalization performances with stronger noises. The observation images in distracting environments are depicted in \autoref{figure:dmc_eval}.}
\label{figure:dcs_test}
\end{figure}

\autoref{figure:dcs_training} and \autoref{figure:dcs_test} show the learning curves of agents in Distracting Control Suite. In Distracting Control Suite, the training environments do not present various styles, unlike Procgen. However, although the SAR agents are not trained with a wide enough range of training environments showing not the best performance in the training phase, they adapt to distracting environments. They show the best performances in three out of four tasks. Here, we apply exponential moving average smoothing only for the training curve with the smoothing coefficient value of $0.99$.

\section{{Curriculum Learning}}
We conduct an experiment on curriculum learning in both Procgen and Distracting Control Suite benchmarks. As addressed by Ko $\&$ Ok\cite{InDA/ExDA}, the timing for the adoption of data augmentation may affect the test performance. 

In Distracting Control Suite environment, while the SAR agents show no better performance with zero noise setting than its baseline SAC~\cite{SAC}, they get improved on several tasks by adopting curriculum learning. The SAR agents also show better performances in several tasks in Procgen with a warm-up stage.

\pagebreak{}
\subsubsection{Procgen.} The SAR agents are trained for 50M timesteps on Procgen with three different start times for applying the adversarial loss: from the beginning, after 10M timesteps of warm-up, and after 25M timesteps of warm-up. The results are averaged over three runs with different seeds.

\begin{table}[h!]
\centering
\caption{The generalization results on Procgen benchmark with different curriculum learning. The best result in the final is bold, and the second result in the final is underlined.}
\begin{tabular}{l||c|cc|cc}
\hline
\multirow{2}{*}{} & from start       & \multicolumn{2}{c|}{after 10M}           & \multicolumn{2}{c}{after 25M}           \\ \cline{2-6} 
                  & at final         & \multicolumn{1}{c|}{at 10M}   & at final & \multicolumn{1}{c|}{at 25M}   & at final \\ 
\hline \hline
                  
\texttt{startpilot}& {27.6±7.9}         & \multicolumn{1}{c|}{27.6±7.9} & \underline{30.7±5.3} & \multicolumn{1}{c|}{21.4±4.1} & \textbf{34.0±13.9} \\ \hline

\texttt{climber}   & \textbf{8.4±0.5}          & \multicolumn{1}{c|}{5.6±2.7}  & \underline{7.1±0.1}  & \multicolumn{1}{c|}{6.3±2.8}  & {6.5±2.6}  \\ \hline

\texttt{jumper}   & \underline{6.0±1.0} & \multicolumn{1}{c|}{6.3±0.6}  & \underline{6.0±1.0}  & \multicolumn{1}{c|}{4.0±2.7}  & \textbf{6.3±2.1}  \\ 
\hline

\texttt{ninja}   & {6.0±1.0} &
\multicolumn{1}{c|}{5.0±2.7} & \underline{6.7±0.6} & \multicolumn{1}{c|}{6.3±0.6}  & \textbf{8.3±1.2} \\ 
\hline

\texttt{coinrun}   & \underline{7.7±0.6}          & \multicolumn{1}{c|}{7.7±0.6}  & 7.0±1.0  & \multicolumn{1}{c|}{8.3±1.2}  & \textbf{8.3±0.6}  \\ \hline

\texttt{maze}   & \underline{6.3±2.1} 
& \multicolumn{1}{c|}{6.3±0.6}  & {5.0±2.0}  & \multicolumn{1}{c|}{{7.0±1.0}}  & \textbf{7.3±3.1}  \\ 
\hline

\texttt{bigfish}   & {6.6±1.0} &
\multicolumn{1}{c|}{1.6±0.6}  & \textbf{12.3±5.4}  & \multicolumn{1}{c|}{3.3±3.5}  & \underline{8.9±0.1}  \\ 
\hline

\texttt{dodgeball}   & \underline{2.1±1.8} &
\multicolumn{1}{c|}{0.5±0.3}  & \underline{2.1±2.5}  & \multicolumn{1}{c|}{1.4±0.7}  & \textbf{3.4±2.3}  \\ 
\hline

\end{tabular}
\end{table}

\subsubsection{Distracting Control Suite.}
The generalization results in Distracting Control Suite with curriculum learning. The SAR agents have trained 500k timesteps with applying the adversarial loss after 300k of warm-up. The results are averaged over three runs with different seeds.

\begin{table}[h!]
\centering
\setlength{\tabcolsep}{5pt}
\caption{The generalization results on Distracting Control Suite with curriculum learning. The best results are in bold.}
\label{tab:curr_dmc}
\begin{tabular}{l|c||c|c}
    \hline
    \multicolumn{2}{c||}{}                                    & SAR from start  & SAR after 300k \\ \hline\hline
    \multirow{3}{*}{{\begin{tabular}[l]{@{}l@{}}\texttt{walker}\\ \texttt{:walk}\end{tabular}}} & zero noise      & 325±57 & \textbf{420±78}    \\
                                               & \texttt{moderate}                                   & \textbf{139±19} & 113±23    \\
                                               & \texttt{hard}                                       & \textbf{112±15} & 88±20    \\ \hline
    \multirow{3}{*}{{\begin{tabular}[l]{@{}l@{}}\texttt{cartpole}\\ \texttt{:balance}\end{tabular}}} & zero noise  & 990±5 & \textbf{996±2}  \\
                                               & \texttt{moderate}                                                  & \textbf{266±26} & 249±10 \\
                                               & \texttt{hard}                                                      & \textbf{261±17} & 241±16 \\ \hline
    \multirow{3}{*}{{\begin{tabular}[l]{@{}l@{}}\texttt{reacher}\\ \texttt{:easy}\end{tabular}}} & zero noise       & 177±51 & \textbf{211±33} \\
                                               & \texttt{moderate}                                                        & \textbf{98±13} & 85±19  \\
                                               & \texttt{hard}                                                            & \textbf{93±10} & 77±14  \\ \hline
    \multirow{3}{*}{{\begin{tabular}[l]{@{}l@{}}\texttt{cheetah}\\ \texttt{:run}\end{tabular}}} & zero noise                                    & \textbf{304±80} & 277±47 \\
                                                & \texttt{moderate}                                                                & 49±11 & \textbf{57±16}  \\
                                                & \texttt{hard}                                                                    & 46±13 & \textbf{53±13}  \\ \hline
    \end{tabular}%
\end{table}

\section{Learned Feature Analysis}

\subsubsection{Image reconstruction from embedded features.}
\begin{figure}[h]
\centering
\includegraphics[width=0.75\textwidth]{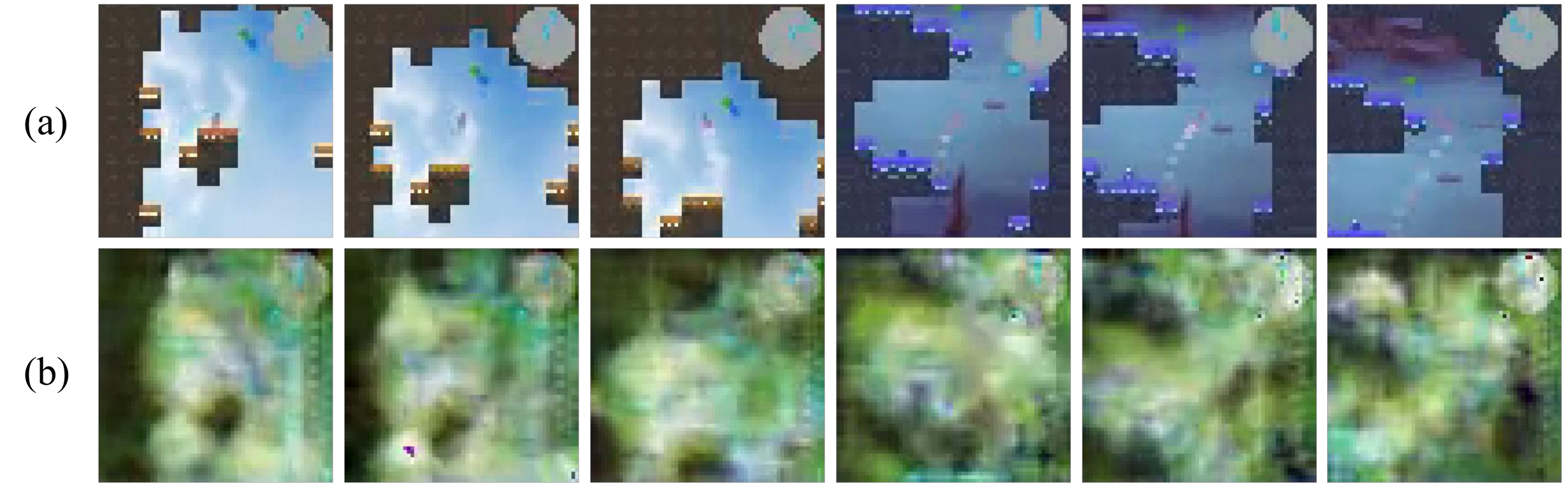}
\caption{(a) Images from two episodes of a trained SAR agent and (b) reconstructed images with the representation features from the agent with \texttt{jumper} in Procgen.}
\label{figure:video}
\end{figure}

\autoref{figure:video} displays three consecutive original frames and their corresponding reconstructed images with embeddings from a trained SAR agent on two different levels of \texttt{jumper} in Procgen. This well describes that the SAR agent is extracting style-agnostic representation features while preserving important elements from the images.

\subsubsection{t-SNE Analysis.}
\begin{figure}[h]
\centering
\includegraphics[width=0.75\textwidth]{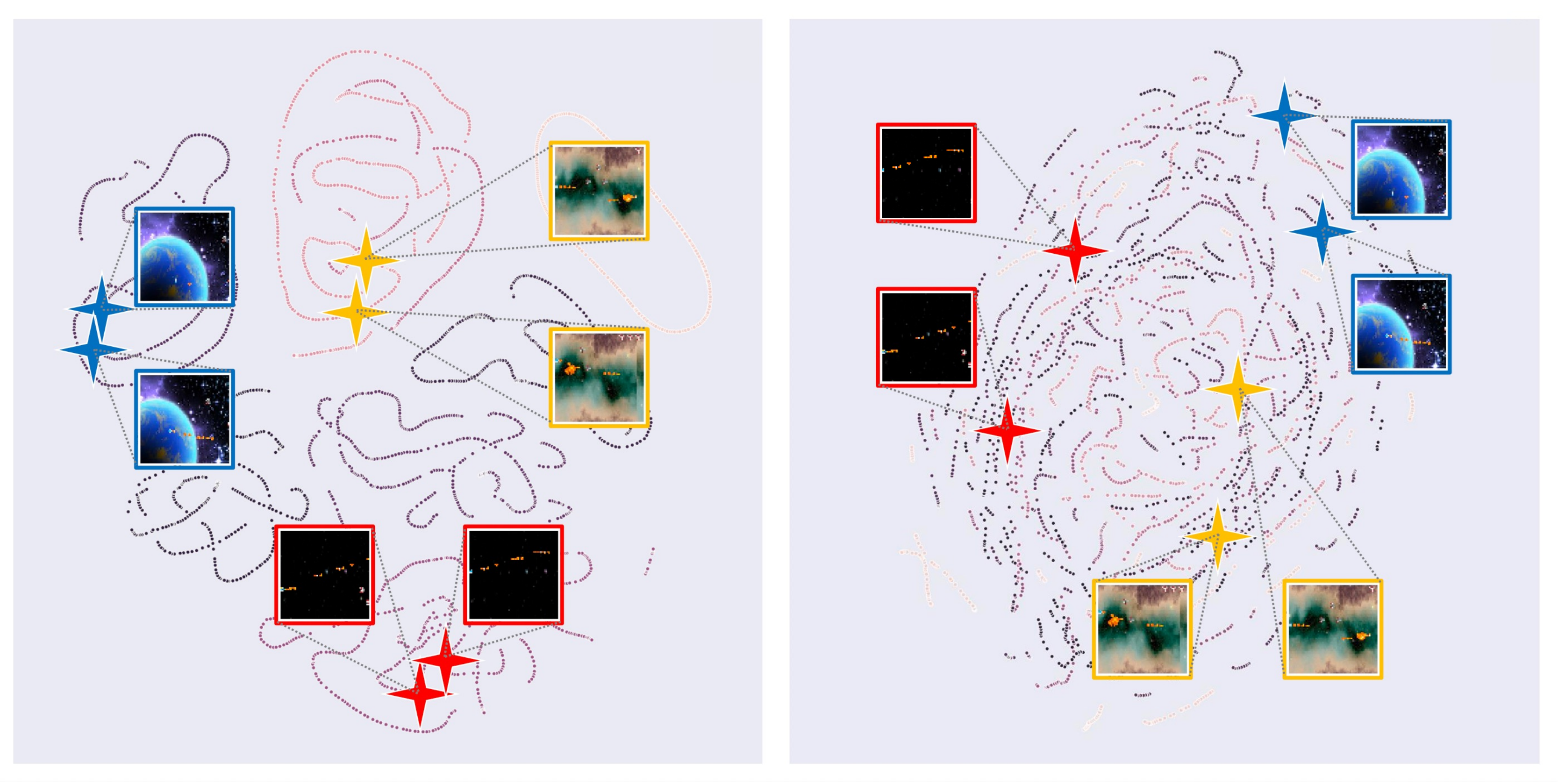}
\caption{t-SNE results of (left) PPO and (right) SAR with the \texttt{starpilot} in Procgen. Example image pairs within 15 frames apart from 3 different levels are marked.}
\label{figure:t_sne}
\end{figure}

\autoref{figure:t_sne} presents the t-SNE results from embeddings extracted from trained the SAR and PPO agents. While the representation features extracted from the PPO agent are well grouped concerning the styles of environments, those from the SAR agent are scattered without a certain pattern. The average distance between all the PPO sample pairs, where a sample is composed of two features five frames apart, is 1.21, while that of SAR sample pairs is 3.41. Also, with sample pairs composed of features 15 frames apart, that of PPO is 3.43, while that of SAR is 9.30. 

